\documentclass[review]{elsarticle}

\usepackage{graphicx}
\usepackage{wrapfig,lipsum,booktabs}
\usepackage{lineno,hyperref}
\usepackage{lscape}
\usepackage[]{algorithm2e}
\modulolinenumbers[5]

\journal{Journal of \LaTeX\ Templates}

\begin{document}

\begin{frontmatter}

\title{Reducing training requirements through evolutionary based dimension reduction and subject transfer}


\author[a1,a2]{Adham Atyabi\corref{mycorrespondingauthor}}
\ead{Adham.Atyabi@Flinders.edu.au;A.Atyabi@salford.ac.uk}
\author[a1,a3]{Martin Luerssen}
\author[a4]{Sean P. Fitzgibbon}
\author[a1]{Trent Lewis}
\author[a1,a5]{David M.W. Powers}
\cortext[mycorrespondingauthor]{Corresponding author}

\address[a1]{Center for Knowledge and Interaction Technologies, School of Computer Science, Engineering, and Mathematics, Flinders University of South Australia, Australia}
\address[a2]{Centre for Autonomous systems and Advance Robotics, University of Salford, United Kingdom}
\address[a3]{Medical Device Research Institute, Flinders University of South Australia, Australia}
\address[a4]{Centre for Functional MRI of the Brain (FMRIB Centre), University of Oxford, United Kingdom}
\address[a5]{Beijing Municipal Lab for Multimedia \& Intelligent Software, Beijing University of Technology, Beijing, China}
\begin{abstract}
Training Brain Computer Interface (BCI) systems to understand the intention of a subject through Electroencephalogram (EEG) data currently requires multiple training sessions with a subject in order to develop the necessary expertise to distinguish signals for different tasks. Conventionally the task of training the subject is done by introducing a training and calibration stage during which some feedback is presented to the subject. This training session can take several hours which is not appropriate for on-line EEG-based BCI systems. An alternative approach is to use previous recording sessions of the same person or some other subjects that performed the same tasks (subject transfer) for training the classifiers. The main aim of this study is to generate a methodology that allows the use of data from other subjects while reducing the dimensions of the data. The study investigates several possibilities for reducing the necessary training and calibration period in subjects and the classifiers and addresses the impact of i) evolutionary subject transfer and ii) adapting previously trained methods (retraining) using other subjects data. Our results suggest reduction to 40\% of target subject data is sufficient for training the classifier. Our results also indicate the superiority of the approaches that incorporated evolutionary subject transfer and highlights the feasibility of adapting a system trained on other subjects.
\end{abstract}

\begin{keyword}
Particle Swarm Optimization \sep Dimension Reduction \sep Brain Computer Interface \sep Subject Transfer 
\end{keyword}

\end{frontmatter}


\section{Introduction}
Interaction between brain and computer is traditionally mediated by the peripheral nervous system using motor nerves and muscles to convey information/instruction to the computer.  A brain-computer interface (BCI) is a device that bypasses the peripheral nervous system and permits the direct transfer of information from the brain to a computer.  A BCI is essentially a device that maps an input onto an output.  The input is the activity of the brain as measured by some sensing device and the output is the control of a device such as cursor movement, spelling device, wheel chair, or prostheses. 

The electroencephalogram (EEG) measures local field potentials that have propagated from the cortex to the scalp.  EEG is an attractive signal measurement tool for a BCI as it is both inexpensive, non-invasive, and has a high temporal resolution. The control signal in a BCI is the pattern of brain activity that reflects the intention/desire of the user to effect control over the BCI.  The primary objective of a BCI is to accurately extract and interpret this control signal.  Numerous EEG phenomena have been exploited as control signals for non-invasive BCIs, with one of the most prominent being sensorimotor rhythms (SMR). SMR are oscillatory EEG activity generated in cortical sensorimotor areas.  SMR can be modulated by preparation for movement, execution of movement, or by imagination of movement \cite{Pfurtscheller1996a}.  It is the effect of motor imagination on SMR that has been exploited by numerous BCI research groups.  Specifically, SMR in the $\mu$ (8-12 Hz) and $\beta$ (18-26 Hz) frequency bands have been shown to be effective control signals for BCI operation. 

An emerging trend of recent years has been the use of advanced machine learning algorithms to create BCI's that are optimised to subject specific patterns of brain activity.  Numerous groups have implemented machine learning algorithms to minimise the need for user-training \cite{Pfurtscheller1996a,Blankertz2003a, Flotzinger1994a,  Obermaier2001a, Millan2002a, Roberts2000a, Penny2000a, Garrett2003a, Hassani2014}.  Machine learning based BCI systems typically operate using some variant of following steps:
\begin{enumerate}
\item Calibration: data is recorded whilst the user repeatedly performs some mental task in which the intention of the user is controlled.
\item Training: the BCI uses the calibration data to optimize the weights of its classifier in a way that accurately classifies the predefined intention of the user
\item Feedback: the trained classifiers are applied to new brain activity to predict the intention of the user
\end{enumerate}

The volume of data obtained when acquiring EEG from many channels with high sample rates and repeated trials can be large.  The dimension of EEG data is the number of channels $\times$ the number of features (e.g. samples, spectral power, etc.) in a single trial and is usually very high.   The high-dimension of EEG data means that many trials are required for successful classification otherwise the problem is underdetermined.  The process of acquiring many calibration trials can be both time-consuming and laborious for the user and may span multiple sessions and/or days.  In addition, the process of training the classifier on large volumes of high-dimensional data can be computationally expensive and also very time-consuming, often spanning days. These combined time constraints limit the real-world and real-time utility of BCI.  Ideally a BCI would be calibrated on a small amount of data acquired in a very short time, and the classifiers trained rapidly to produce a functional and practical BCI.

Dimension reduction (DR) can be utilized to reduce the volume of data prior to classification, and as a consequence reduce computational demands and processing time. The challenge of DR is to maximise compression whilst minimising loss of information and maintaining performance.  DR can be performed on channels, features, or both. Decomposition-based methods (e.g. principal components analysis) are popular choices for feature and electrode reduction however they use a procedure that mostly isolates a set of features or electrodes that satisfy a certain statistical criterion without considering their actual impact on the classifier's learning rule. Evolutionary methods such as Genetic Algorithm (GA) and Particle Swarm Optimization (PSO) are alternative approaches that can be utilized for dimension reduction and are not subject to this limitation. These approaches evolve their population toward a subset of features and/or channels that achieve the best classification performance.  However these evolutionary methods are very computationally intensive which further exacerbates the time constraints of the BCI system \cite{EA12,EA1}. An evolutionary dimension reduction approach, called PSO-DR, that allows simultaneous feature and electrode reduction was proposed in \cite{WCCIPSO}. In this approach, PSO is utilized to identify a subset of feature and electrode indexes that best represent the performed task and offers better generalizability. In order to address the computationally intensiveness of the PSO-DR, subject transfer, which is the use of previously recorded samples originating from other subjects \cite{STx1,STx2,STx3,STx4}, is utilized to pre-identify the subset of feature and electrode indexes by the PSO-DR \cite{WCCICat, FRERPSOJournal}.

The aim of this study is to combine subject transfer with evolutionary dimension reduction (PSO-DR) to reduce the dimensionality of the data and the number of training trials required for classification. Advantages of this methodology are:
\begin{itemize} 
\item Extra training samples originating from other subjects can be prerecorded and preprocessed off-line.
\item Minimum dependency on training trials from the target subject (the subject that performs the tasks in on-line).
\item Low dimensionality of the data as a result of evolutionary dimension reduction that reduces the required classifier training time. 
\end{itemize}

\section{Analysis and Methods}

\subsection{Particle Swarm Optimization (PSO)}
\label{history}
Particle Swarm Optimization (PSO)  is an evolutionary approach inspired by animals' social behaviors. PSO accommodates search through generating, evaluating and updating members of population called particles that each represent a possible solution. Particles in the population (swarm) are known by their position in the search space (called \emph{X}) and their velocity (called \emph{V}) and remember the best solution found by them (called personal best or PBest) and their neighboring particles or the entire swarm (called global best or GBest). In PSO particles in the swarm update their velocity and their position in the search space using the PBest and the GBest as reference points. The equations for updating the velocity and particles' positions are as follow:
\begin{equation}
\label{eq:basic pso velocity equation}
\begin{array}{c}
V_{i,j}(t)= w V_{i,j}(t-1)+C_{i,j} + S_{i,j}\\
C_{i,j} = c_1 r_{1,j} \times (PBest_{i,j}(t-1)- x_{i,j}(t-1))\\
S_{i,j} = c_2 r_{2,j} \times (GBest_{i,j}(t-1)- x_{i,j}(t-1))
\end{array}
\end{equation}
In Eq. \ref{eq:basic pso velocity equation}, $V_{i,j}(t)$ represents the velocity in iteration $t$. \emph{i} and \emph{j} represent the particles index and the dimension in the search space respectively. $c_1$ and $c_2$ are the acceleration coefficients utilized to control the effect of the cognitive ($C_{i,j}$) and social ($S_{i,j}$) components respectively.  $r_{1,j}$ and $r_{2,j}$ are random values in the range of [0,1] and $w$ is the inertia weight that controls the influence of the last velocity in the updated version.

Various mechanisms have been suggested for adjusting the parameters in Equation \ref{eq:basic pso velocity equation}, including Linearly Decreasing Inertia Weight (LDIW) \cite{Pasupuleti}, Time Varying Inertia Weight (TVIW) \cite{Pasupuleti,Vesterstrom2,Ratnaweera}, Linearly Decreasing Acceleration Coefficient (LDAC) \cite{Pasupuleti,Ratnaweera,Chang,Watts}, Time Varying Acceleration Coefficient (TVAC) \cite{Pasupuleti,Vesterstrom2,Ratnaweera}, Random Inertia Weight (RANDIW) \cite{Ratnaweera}, Fix Inertia Weight (FIW), Random Acceleration Coefficients (RANDAC), and Fix Acceleration Coefficients (FAC) \cite{Ratnaweera}. Among the proposed modifications in several studies, LDIW and FixAC are utilized in here. Equation \ref{W} represents the LDIW formulation.
\begin{equation}
\label{W}
w=(w_1 - w_2) \times \frac{(maxiter - t)}{maxiter}+ w_2
\end{equation}
where, $w_1$ and $w_2$ are the initial and final inertia weight, {\em t} is the current iteration, and \emph{maxiter} is the termination iteration. The equation for updating the particles are as follows:

\begin{equation}
\label{eq:basic pso particle equation}
x_{i,j}(t)= x_{i,j}(t-1) + V_{i,j}(t)
\end{equation}
$PBest_{i,j}$  and $GBest_{i,j}$ represent the best solution found by individual particles and the best overall solution found by the swarm and can be updated using Equations \ref{P} and \ref{G}. In the equations, \emph{f} represents the fitness function utilized to assess the feasibility of the particle (\emph{x}), personal best solution (\emph{PBest}) or global best solution (\emph{GBest}).

\begin{equation}
\label{P}
PBest_{i}(t) =
\left\{\begin{tabular}{l}
$PBest_{i}(t - 1),$~ if ~$f(x_{i}(t)) \preceq f(PBest_{i}(t - 1))$\\
$x_{i}(t),$~ otherwise~\\
\end{tabular}
\right.
\end{equation}

\begin{equation}
\label{G}
GBest(t) = argmin\left\{f(PBest_{1}(t)),...,f(PBest_{n}(t))\}\right.
\end{equation}

A pseudocode listing of the PSO approach is presented in Algorithm. \ref{FRPSO}. Comprehensive reviews of PSO can be found in \cite{Bookchapter1,Nadhir}. Furthermore, a review of existing studies utilizing evolutionary approaches in EEG-based brain computer interfacing studies can be found in \cite{Bookchapter2}.

\begin{algorithm}[H]
{\bf Initialization:} Randomly initialize the population.\\
{\bf Initial Evaluation:} Evaluate all members of the population using the fitness function \emph{f}.\\
 \While{The maximum iteration is not reached or the best member of the population (\emph{GBest}) is performing below the highest expected performance}{
  1. \emph{\bf Update the population}: Update the velocity in each particle using equation \ref{eq:basic pso velocity equation} and update the particle by applying the new velocity to equation \ref{eq:basic pso particle equation}.\\
  2. \emph{\bf Evaluation}: Evaluate all members of the population using the fitness function \emph{f}.\\
  3. \emph{\bf Updating the best findings}: Update \emph{PBest} and \emph{GBest} using equations \ref{P} and \ref{G}.\\
 }
 \caption{Basic PSO Pseudocode}
 \label{FRPSO}
\end{algorithm}


\subsection{PSO Dimension Reduction (PSO-DR)}

In the study of EEG, dimension reduction can be applied on both feature and/or channel dimensions. Atyabi et al. \cite{WCCIEA} investigated the impact of evolutionary approaches such as Genetic Algorithm, Random Search, and PSO for feature and channel reduction among which the PSO-based feature reduction approach showed better overall generalizability. Atyabi et al. \cite{WCCIPSO} proposed 99\% reduction through simultaneous reduction of feature and channel sets using a PSO-based approach with two layer swarm structure called PSO-DR. In the study, PSO-DR showed better overall performance in comparison to variety of approaches including 2 variations of Single Value Decomposition (i.e., SVD-based electrode reduction and SVD-based decomposition) and PSO-based electrode reduction.

\subsection{The structural design of the PSO-DR}

The structural design of the PSO-DR can be expressed within a two layer swarm notation containing a population of possible solutions referred to as \emph{Masks} in the first layer. The second layer focuses on the \emph{masks} by treating them as sub-swarms. Each \emph{mask} have a small memory containing an \emph{Electrode Vector} (denoted as \emph{ELV}) representing the selected electrode indexes and a set of particles bundled in a matrix notation called \emph{Feature Set Matrix} (denoted as \emph{FSM}). Each particle in the \emph{FSM} matrix (rows of the matrix) represents a set of indexes from feature-space that are down selected by PSO for the designated electrode in the \emph{ELV}. 

Given an EEG dataset with \emph{N} electrodes and \emph{K} feature points in each epoch of each electrode, \emph{ELV} vector in each \emph{mask} of the swarm contain \emph{n} out of \emph{N} electrodes while the \emph{FSM} contains $n \times k$ out of $N \times K$ possible features (indexes). 

\begin{figure}
\begin{center}
\includegraphics[height=7cm]{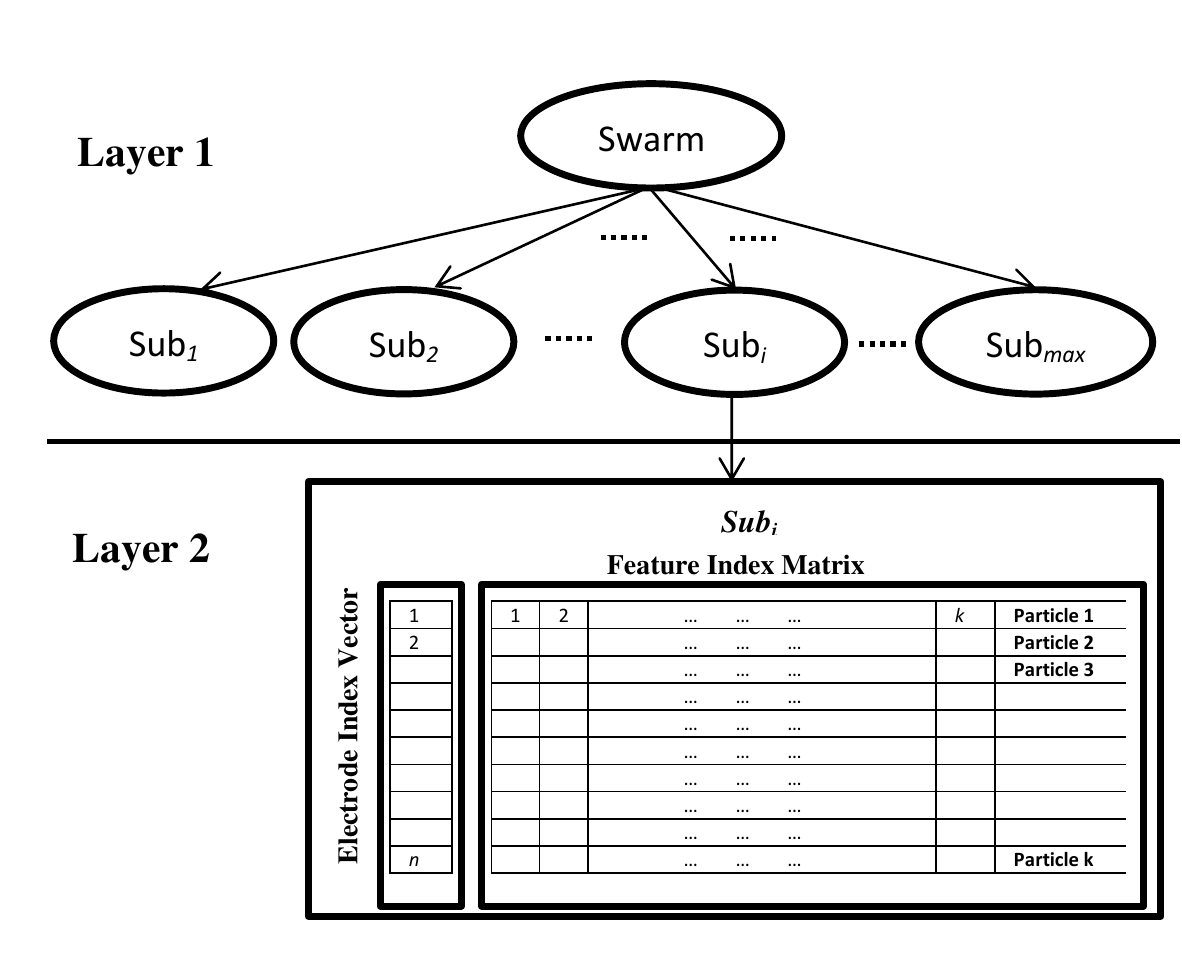}
\end{center}
\caption{A diagram representing the PSO-based mechanism employed for simultaneous feature and electrode reduction. In the figure, \emph{Sub} represent sub-swarm. Each sub-swarm contains \emph{n} out of \emph{N} electrode indexes and \emph{k} out of \emph{K} feature point indexes for each of the selected electrodes. \emph{max} index in $Sub_{max}$ represent the swarm's population size.}
\label{2dSwarm}
\end{figure}

In addition, each \emph{mask} contains a velocity vector $V_{i}$ for $i \in [1 ... n]$ representing the previous velocity for each particle in the feature set matrix. Adapting this notation for Eq. \ref{eq:basic pso particle equation}, each particle in the sub-swarm can be denoted as $X_{i,j}$ for $i \in [1 ... n]$ and $j \in [1 ... k]$. Figure \ref{2dSwarm} provides a simple representation of the two layer swarm utilized for simultaneous feature and electrode reduction.

In the swarm, \emph{PBest$_{i}$} for $i \in [1 ... max]$ represent the personal best findings of the masks. This is equivalent to a copy of the matrix of the masks that received the highest classification performance.

Therefore, a member of the swarm (sub-swarm) called \emph{$Sub_{i}$} represents a mask with \emph{n} electrodes and $n \times k$ features. In each iteration, the sub-swarms share their best found {mask} (the mask with the highest classification performance) called \emph{PBest}. \emph{GBest} is considered as the global best and represents the best \emph{PBest}. 
 
\begin{figure}
\begin{center}
\includegraphics[width=3.3in]{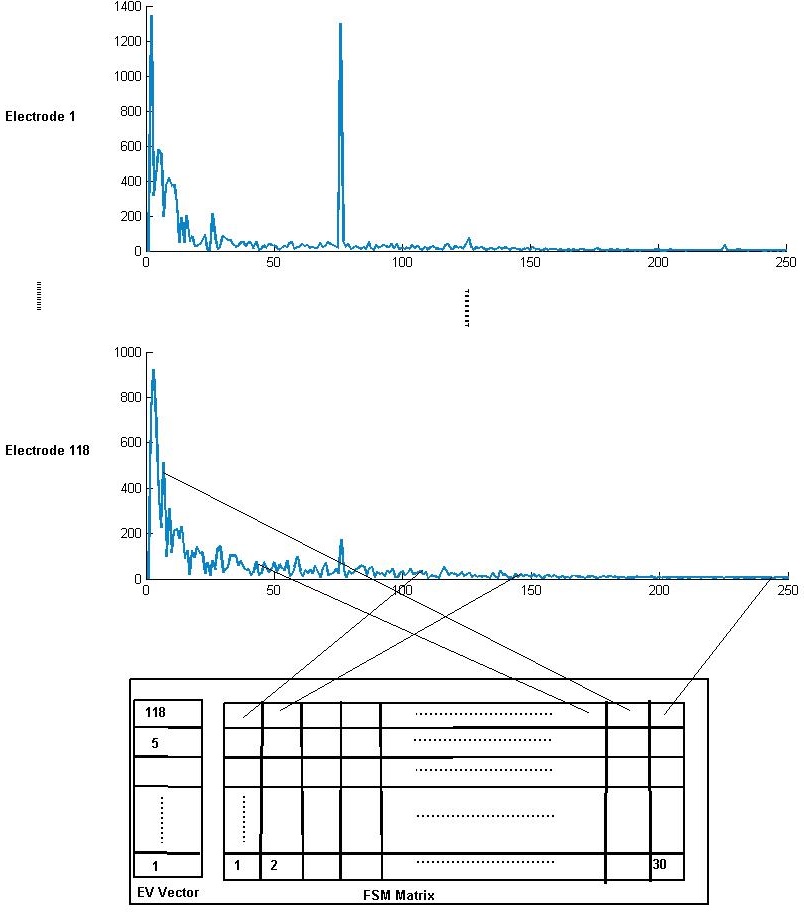}
\end{center}
\caption{A sample of the PSO-DR mask created for a dataset with \emph{N=118} electrode and \emph{K=250} feature indexes. \emph{n} and \emph{k} are set to 10 and 30 respectively.}
\label{PsoParadigmx}
\end{figure}

Table \ref{PSOParadigmSetting} provides more details about the settings of PSO-DR and Fig. \ref{PsoParadigmx} illustrates an example of the utilized \emph{mask} in PSO-DR.

\begin{table}
\center
\begin{scriptsize}
\caption{The initial settings of the PSO-DR}
\begin{tabular}{c|cc} 
\hline
&{\bf Details}&{\bf Dimension}\\\hline
EEG dataset&\emph{N} Electrodes, \emph{K} Feature Points&\emph{N$\times$K}\\\hline
Each Mask& EV vector and FSM Matrix&\\\hline
EV vector&\emph{n} out of \emph{N} electrodes&\emph{n$\times$1} electrode indexes\\\hline
Each row of FSM Matrix&\emph{k} out of \emph{K} feature points&\emph{1$\times$k} feature indexes\\\hline
FSM Matrix&\emph{n$\times$k} out of \emph{N$\times$K}&\emph{n$\times$k} indexes\\\hline
\end{tabular}
\label{PSOParadigmSetting}
\end{scriptsize}
\end{table}

\subsection{Simultaneous feature \& electrode reduction with PSO-DR}

The proposed PSO-DR is utilized within a method of two stages. The first stage includes using PSO to extract a subset of indexes (\emph{mask}), representing \emph{n} out of \emph{N} electrodes and \emph{k} out of \emph{K} features for each electrode (\emph{n=10} and \emph{k=30} in here). This procedure is applied to several fragments of the EEG data in a 10 repetitions of 20 fold cross validation (10 $\times$ 20 CV) manner using a training and validation sets representing 90\% and 5\% of the data resulting in 200 unique masks. These masks are reassessed on unseen testing sets that represent the remaining 5\% of the data within each fold as the final step.  In this stage, the classification performance achieved from a sigmoid Extreme Learning Machine (ELM) \cite{ELM5, ELM4, ELM6} with 80 nodes is utilized as the fitness criteria. The choice of using sigmoid ELM as the classifier is made based on its fast learning capability and its time efficiency. Previous experiments indicated a lack of generalizability for the generated masks as they performed poorly with the unseen testing set \cite{WCCIPSO}. The PSO-DR uses Linearly Decreasing Inertia Weight (LDIW) with $w_{1}=0.2$ and $w_{2}=1$ and fixed acceleration coefficients with $c_{1}=0.5$ and $c_{2}=2.5$\footnote{Other parameter settings such as using fix value for \emph{w} (w=0.729844) and  $c_{1}=c_{2}=0.5$ were also tested but the parameter setting utilized in here showed more consistent performance.}.

\begin{figure}
\begin{center}
\includegraphics[width=4in]{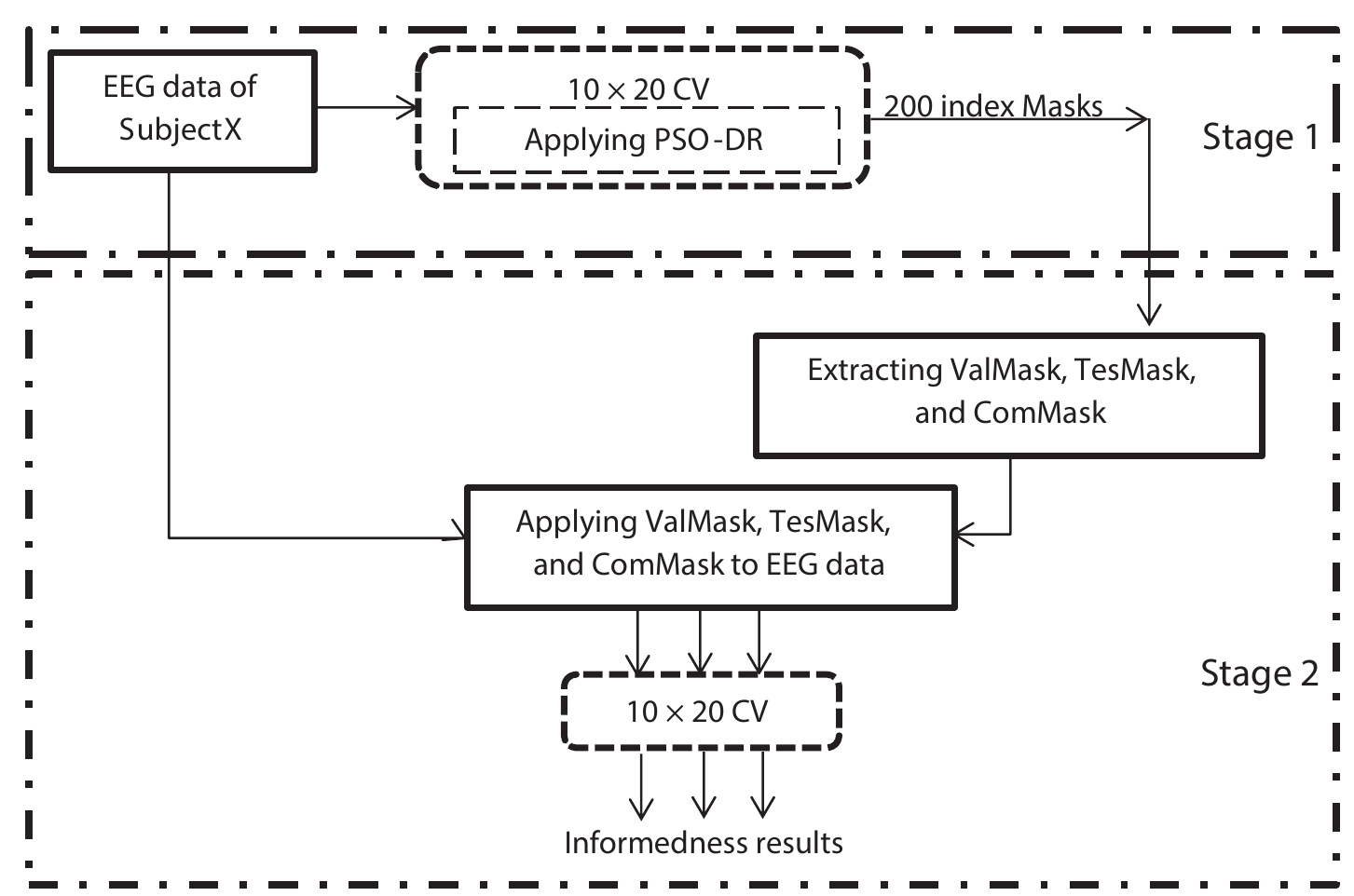}
\end{center}
\caption{Diagram demonstrating the application of PSO-DR on a single subjects as in \cite{WCCIPSO}.}
\label{ApplyPSO}
\end{figure}

This issue has been resolved \cite{WCCIPSO} by extracting three new masks representing the best performing masks on validation and testing sets (BestMask) and the mask containing the most commonly selected indexes from the groups of masks generated in stage 1 (ComMask). The potential of these new masks are assessed using a new 10 $\times$ 20 CV. Figure \ref{ApplyPSO} provides a graphical representation of the procedures performed in the two stages in \cite{WCCIPSO}.

\begin{figure}
\begin{center}
\includegraphics[width=8cm]{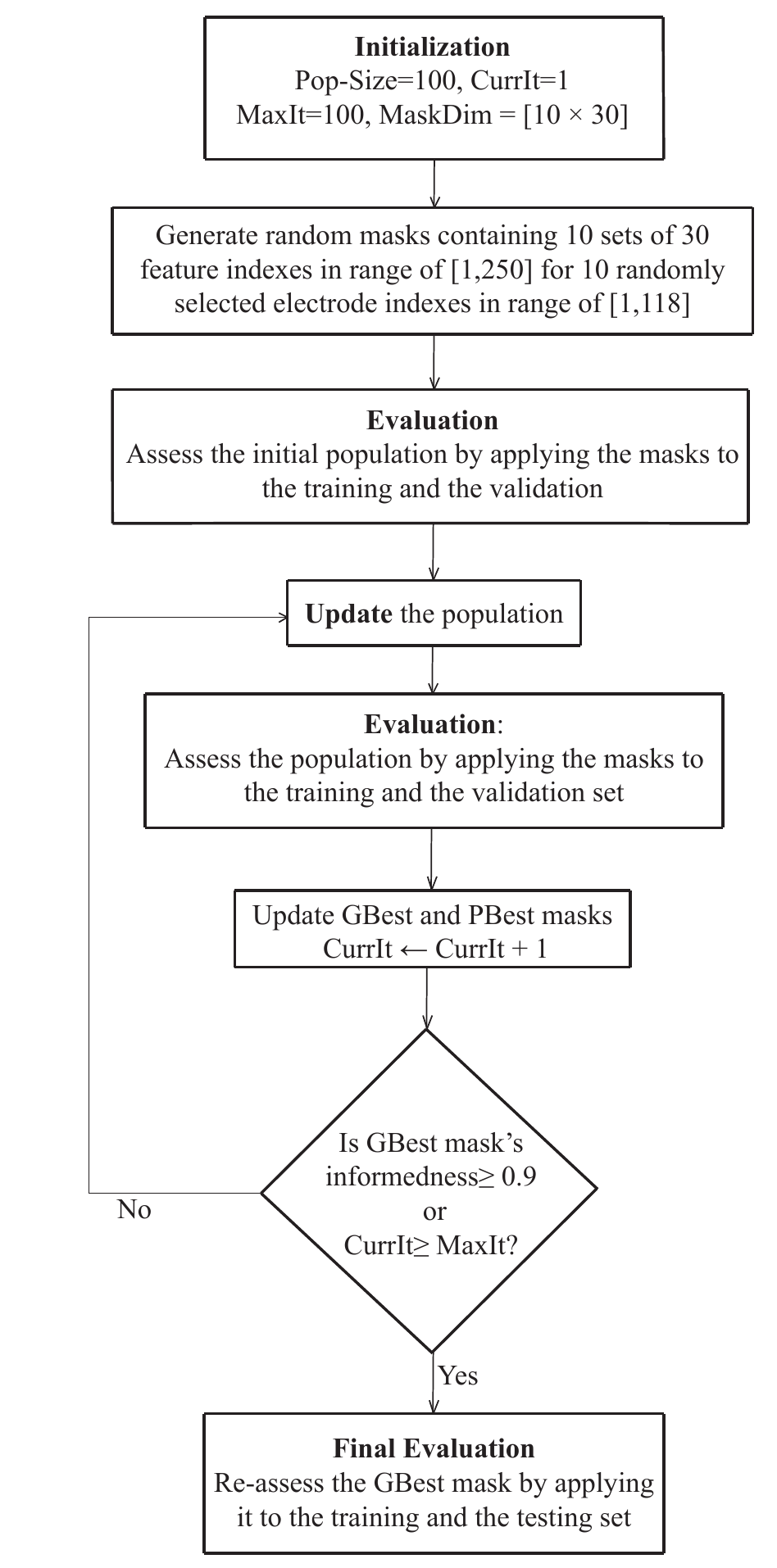}
\end{center}
\caption{Flowchart of the PSO-DR approach applied to an EEG data with 118 electrodes, 0.5s epochs with 1000Hz sample rate. CurrIt, MaxIt, Pop-Size, and MaskDim represent Current Iteration, Maximum Iteration, Population size, and Mask dimension respectively. Global and personal best are denoted as GBest and PBest respectively. The classification performance achieved from a sigmoid ELM is utilized as the fitness criteria. The procedure utilized for updating the particles is illustrated in figure \ref{PsoParadigm2}.}
\label{PsoParadigm}
\end{figure}

\begin{figure}
\begin{center}
\includegraphics[width=8.3cm]{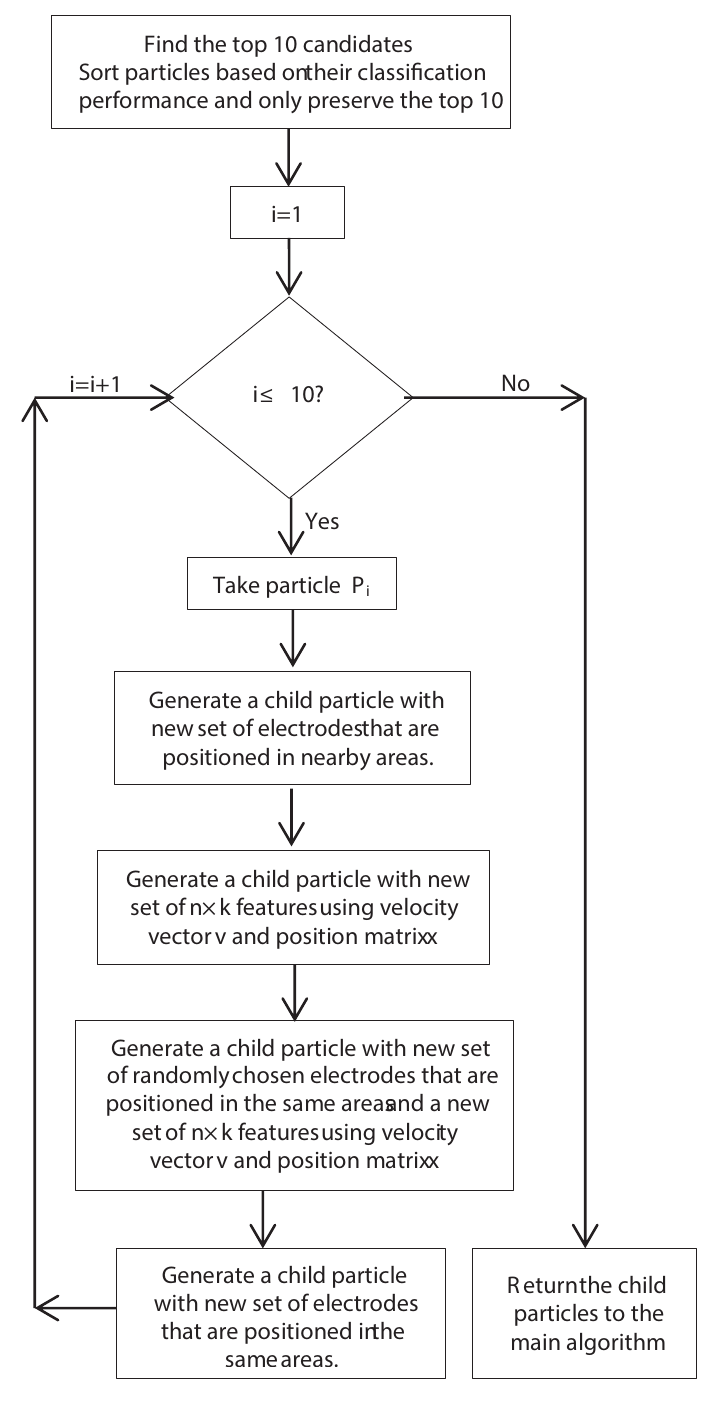}
\end{center}
\caption{Flowchart of particle update procedure utilized in PSO-DR approach.}
\label{PsoParadigm2}
\end{figure}

In the second stage (see Fig. \ref{ApplyPSO}) the sigmoid ELM in addition to two alternative learners (Polynomial SVM and a modified Perceptron with early stopping) are utilized to assess the performance and investigate the transferability of the produced masks from the weak learner (ELM) to alternative learners (SVM and Perceptron). ComMask showed better generalizability than BestMask and the use of stronger learners improved the performance. A flowchart of the PSO-DR approach  and the population update criteria utilized in here are illustrated in Fig. \ref{PsoParadigm} and \ref{PsoParadigm2}.

\subsection{Subject Transfer with PSO-DR}
The potential of the PSO-DR approach is assessed for subject transfer applications in \cite{WCCICat} and \cite{FRERPSOJournal}. These studies introduced two variant frameworks based on the PSO-DR approach proposed in \cite{WCCIPSO} that i) combined sets of masks generated from individual subjects together generating a \emph{Meta Mask} of masks (denoted as Framework 1), and ii) concatenated EEG signals of several subjects together creating a \emph{Super Subject} and applying PSO-DR to generate sets of masks (denoted as Framework 2). The frameworks have subtle differences that are interpreted as addresses \emph{'subject specificity'} (Framework 1) and \emph{'task specificity'} (Framework 2). 

Subject specificity is hypothesized to capture the idea that subjects have different EEG phenotypes and perform differently from certain other subjects, while similarly with subjects with a similar phenotypical signature, and using masks derived from a good subject (prototypically having previous experience relevant to the task) and a moderate subject (prototypically relying on learning strategies rather than experience), the training of a specific mask for a moderate or good subject can be accelerated, and even for a poor subject, greater attention can be paid to successful strategies rather giving equal weight to all the trials, errors and noise associated with a weak subject.

Task specificity is hypothesized to capture information about the distribution of behaviors, strategies or phenotypes in application to the specific task for which data has been collected.

\begin{figure}
\begin{center}
\includegraphics[width=3.0in]{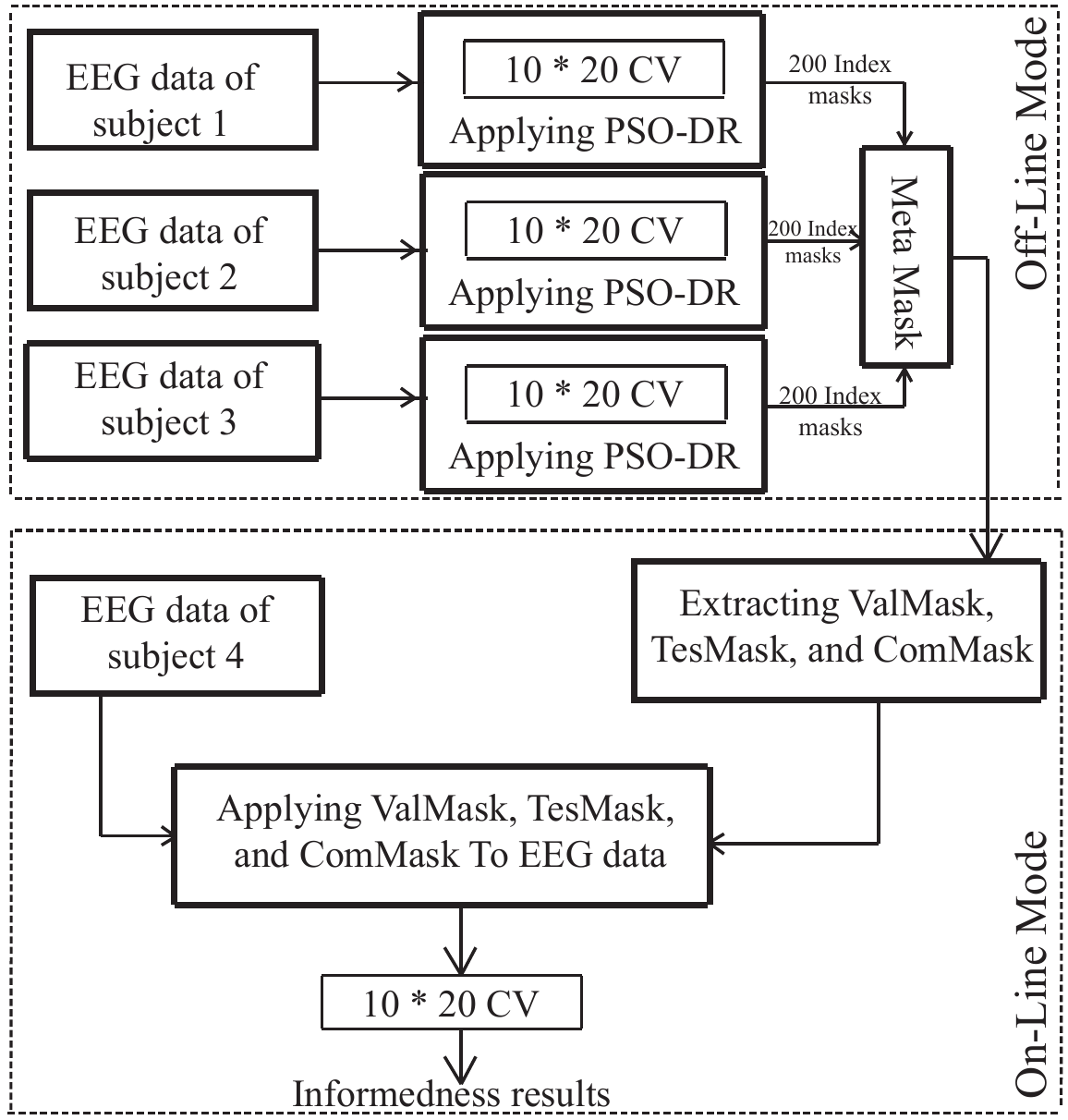}
\end{center}
\caption{A diagram demonstrating the appliance of Framework 1 on a group of 4 subject.}
\label{Framework1}
\end{figure}

\begin{figure}
\begin{center}
\includegraphics[width=3.0in]{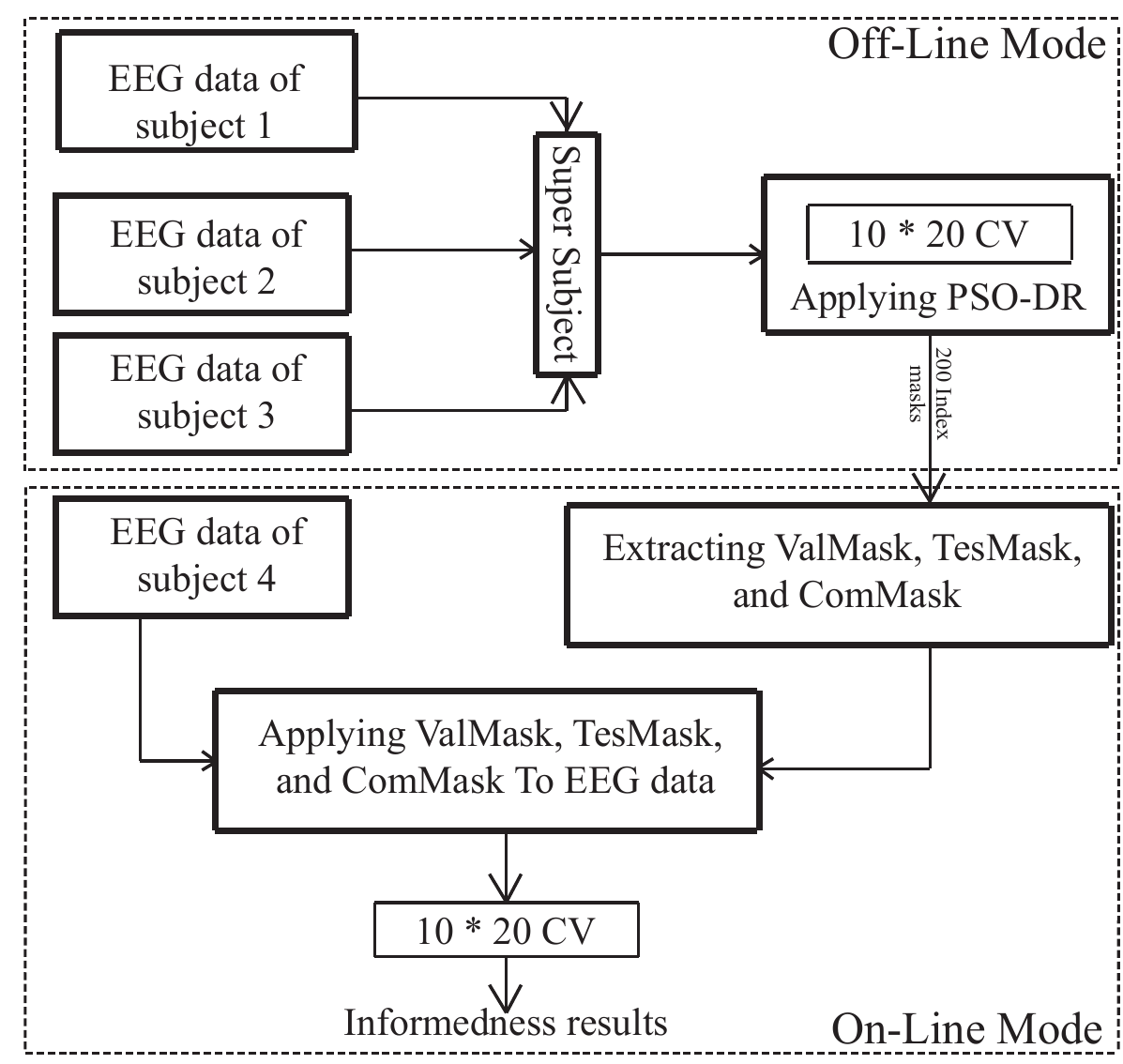}
\end{center}
\caption{A diagram demonstrating the appliance of Framework 2 on a group of 4 subject.}
\label{Framework2}
\end{figure}

The potential of the suggested frameworks is assessed in subject transfer study by introducing the mask containing commonly selected indexes (\emph{ComMask}) resulted from Frameworks 1 and 2 to the EEG data of a new subject (\emph{Target Subject}). Flowcharts of the frameworks utilized in the study featuring a dataset with 4 subjects are illustrated in Fig. \ref{Framework1} and \ref{Framework2}.

Several configurations with different numbers and strengths of subjects involved in mask extraction (either for creating the \emph{Meta Mask} or the \emph{Super Subject}) and their strength in terms of classification performance is investigated \cite{FRERPSOJournal}. The results indicated the feasibility of subject transfer through the use of PSO-DR and highlighted the importance of combining the right subjects. In addition, the combination of \emph{ComMask} and Framework 2 achieved the best overall performance. It is noteworthy that in \cite{WCCICat} and \cite{FRERPSOJournal} the combination of the proposed frameworks and PSO-DR are used to utilize subject transfer via dimension reduction. That is, the samples originating from other subjects performing similar tasks are only utilized to extract PSO-DR masks for reducing the dimensions of EEG data of the target subject and not as extra sources of training samples for the classifiers. All experiments that involved target subject in previous studies (i.e., \cite{WCCICat} and \cite{FRERPSOJournal}) were based on Single Trial principle in which 90\% of the subject's data in each fold of the cross-validation is utilized for training the classifiers and the remaining unseen data are utilized for validation and testing.

Various studies investigated the impact of minimizing the required pre-training sessions and the use of extra samples originating from the same subject (session transfer) or some other subjects performing similar tasks in previous sessions (subject transfer) \cite{STx1,STx2,STx3,STx4,nosubjecttraining6,nosubjecttraining5,nosubjecttraining7,Berlinnosubject, nosubjecttraining4}. 
Krauledat et al., \cite{nosubjecttraining7} explored the possibility of using zero-trial BCI system through creating pre-trained subject specific classifiers. In their study, the classifiers that were pre-trained on several previously recorded EEG sessions of the same subject were utilized in the test data to provide feedback. The classifiers' bias were readjusted in the test data using 20 trials recorded during the calibration stage at that day. Although the study illustrated very interesting results, it is noteworthy that all the subjects that contributed in this study were healthy people with several past experiences with BCI systems that allowed them to perform well under the considered constrains. This issue is further investigated by Fazli et al., in \cite{nosubjecttraining4} in a similar study including 83 subjects with variety of EEG expertise in an effort toward creating a subject independent BCI system. The results depicted a degradation of performance in more expert subjects when zero-training procedure is followed while weaker subjects were benefited from pre-trained classifiers that were optimized based on EEG data of group of other subjects with higher BCI expertise.  
The impact of pre-training sessions on overall performance of subjects is investigated by Blankertz et al. in \cite{Berlinnosubject} in an effort toward an EEG based BCI system with no pre-training session requisite in which the expert/strong subjects showed better performance. 

\subsection{What is New?}
The following study  investigates the possibility of reducing the required training sessions through incorporating data from groups of subjects that performed the same tasks in off-line mode. In addition, the study assesses the impact of the proposed PSO-based subject transfer in \cite{FRERPSOJournal} for reducing the number of required samples that are originating from the target subject for training the classifiers. Such a combination is favorable in on-line EEG based BCI systems due to providing i) extra training samples originating from other subjects recorded and pre-processed in an off-line mode, ii) having minimum dependency to training samples that are originating from the target subject, and iii) solving the dimensionality problem of the EEG data by the application of the PSO-DR which also minimizes the amount of time required for training the classifiers. In order to identify such combination of methods, a set of experiments are designed intending to clarify points such as:
\begin{itemize}
\item the minimum amount of EEG data (originated from a target subject) required for training the classifiers aiming to move from Single Trial scheme toward Zero Trial scheme.
\item the impact of the proposed PSO-based subject transfer on reducing the amount of target subject's data required for training the classifiers.
\item the feasability of adopting classifiers that are previously trained by other subjects' data. 
\end{itemize}

\section{Dataset and Evaluation}
\label{dataset}
BCI competition III dataset IVa is utilized in this study \cite{BCIcompetition}. The data is recorded from 5 healthy subjects performing 2 class motor imagery tasks (i.e., hand versus foot movement imagination) in 3.5s epochs over 280 trials. To be consistent with other studies with the dataset \cite{WCCITri,WCCICat,WCCIPSO,FRERPSOJournal,Sean}, the 3.5s super-epochs are divided to seven 0.5s sub-epochs with first and last 0.5s sub-epochs being omitted (representing prior and post transition periods respectively). The resulting datasets contain EEG recordings from 118 channels over 1400 trials  (5 $\times$ 280) with 1000Hz sample rate. The detail information about them is presented in table \ref{tab:DetailsOfMI2MI4Dataset}.

The dataset is demeaned and common average referenced in the preprocessing stage and absolute Discrete Fourier Transform is applied for extracting frequency features \footnote{Matlab 2010 service pack 1\rq{}s implementation of DFT that uses Fast Fourier Transform is employed}. A modified Perceptron that accommodates early stopping is utilized as the primary classifier (evaluation stage). All experiments are done in a $10 \times 20$ CV\footnote{10 repetitions of 20 fold cross validation.} and summarized into a contingency table, and consistent with \cite{WCCICat,WCCIPSO,FRERPSOJournal}, Bookmaker Informedness is utilized for assessing the performance in terms of value between -1 and 1 where 0 represents chance level and 1 perfect performance. The sign indicates whether the information is being utilized correctly or incorrectly (perversely). Absolute Informedness is formally defined as the probability of an informed decision being made (rather than guessing) and is discussed in detail in \cite{bookmaker1, bookmaker2, bookmaker3}. To improve the understanding of the results the average of the achieved classification accuracy is also reported in the study. 

\begin{table}[h]
\begin{center}
\begin{scriptsize}
	\caption{Details of datasets utilized in the study  \cite{BCIcompetition}.}
	\label{tab:DetailsOfMI2MI4Dataset}
\centering
\begin{tabular}{p{1.5cm} p{1.3cm}p{0.8cm}p{1.cm} p{1.cm} p{1.5cm}p{1.4cm}} 
\hline 
Dataset&Number of EEG Channels&Task Trials&Sample Rate &Task Duration(s)& Number of Participants& Performed Tasks\\ [0.5ex]
 \hline 
\hline 
BCI Competition III, Dataset IVa&118&280&1000Hz&3.5s&5 subjects AA, AL, AV, AW, AY&right-foot, left-hand\\\hline  
\hline\hline 
\end{tabular}
\end{scriptsize}
\end{center}
\end{table}

\section{Experimental Aims, Design, and Achievements}
\label{ExperimentalDesign}
Based on the way that other subjects' data are incorporated with the target subject's data two sets of experiments are designed. 
\begin{enumerate} 
\item In the first set of experiments, Experiment 1, various portions of target subject's data is concatenated to EEG data of a group of subjects and the resulting data is utilized for training the classifier in the on-line mode. This is similar to the subjects' data mixing approach employed in \cite{nosubjecttraining6} and the results are considered as the baseline performance for other experiments. 
\item In the second set of experiments, Experiment 2, the EEG data of other subjects are first utilized to pre-train the classifiers and later on, a portion of EEG data originating from the target subject is utilized to re-train the classifier. This is similar to the subjects' data mixing strategy employed in \cite{nosubjecttraining7,Berlinnosubject,nosubjecttraining4} and the results are considered as the baseline performance for other experiments. 
\end{enumerate}

\begin{figure*}[h]
\begin{center}
\includegraphics[width=5.6in]{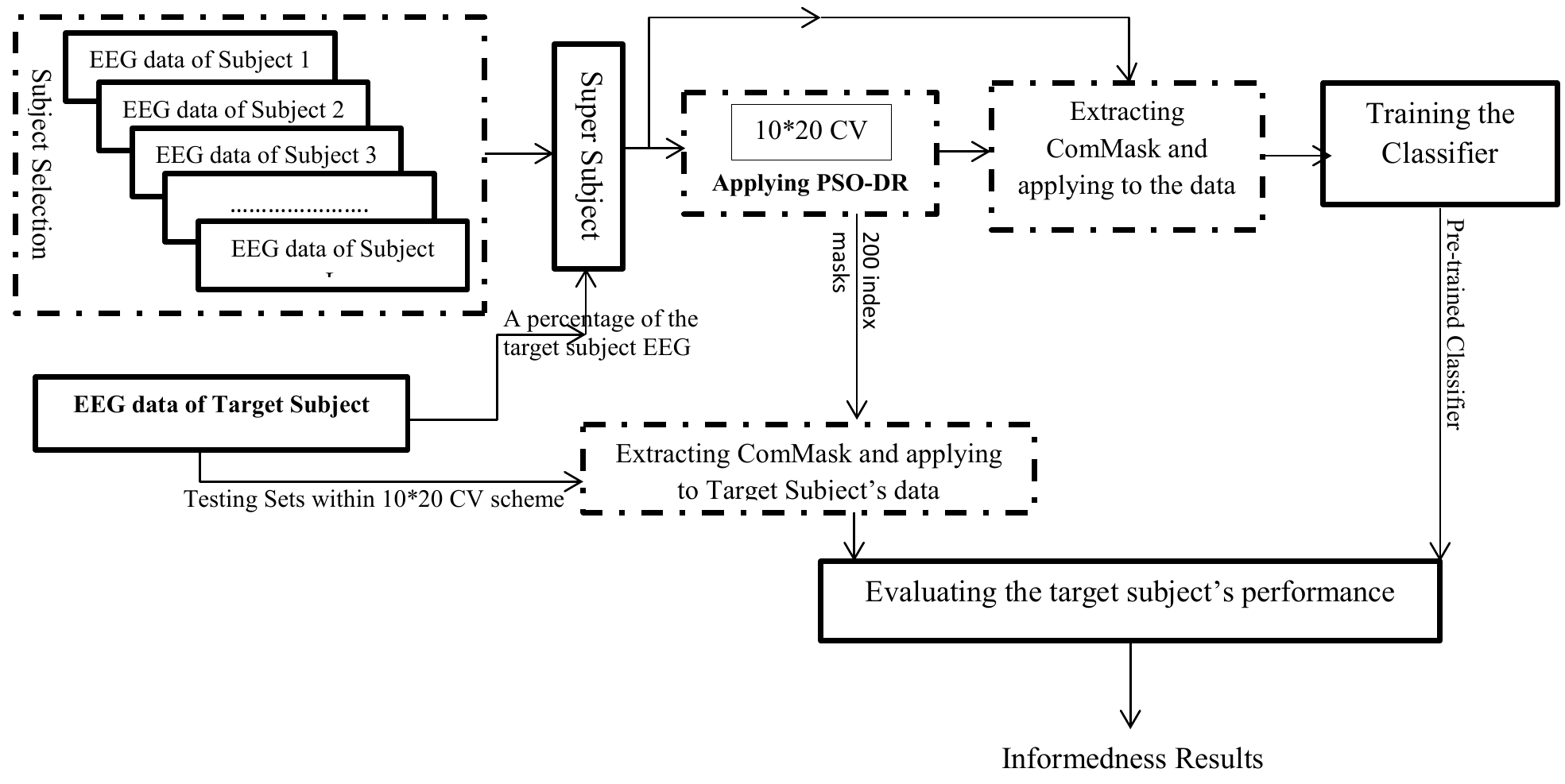}
\end{center}
\caption{A diagram demonstrating the process flow of Experiment 1. Dashed rectangles indicate optional steps that represent sub-experiments (referred to as conditions in the text).}
\label{Exp1}
\end{figure*}

\begin{figure*}[h]
\begin{center}
\includegraphics[width=5.6in]{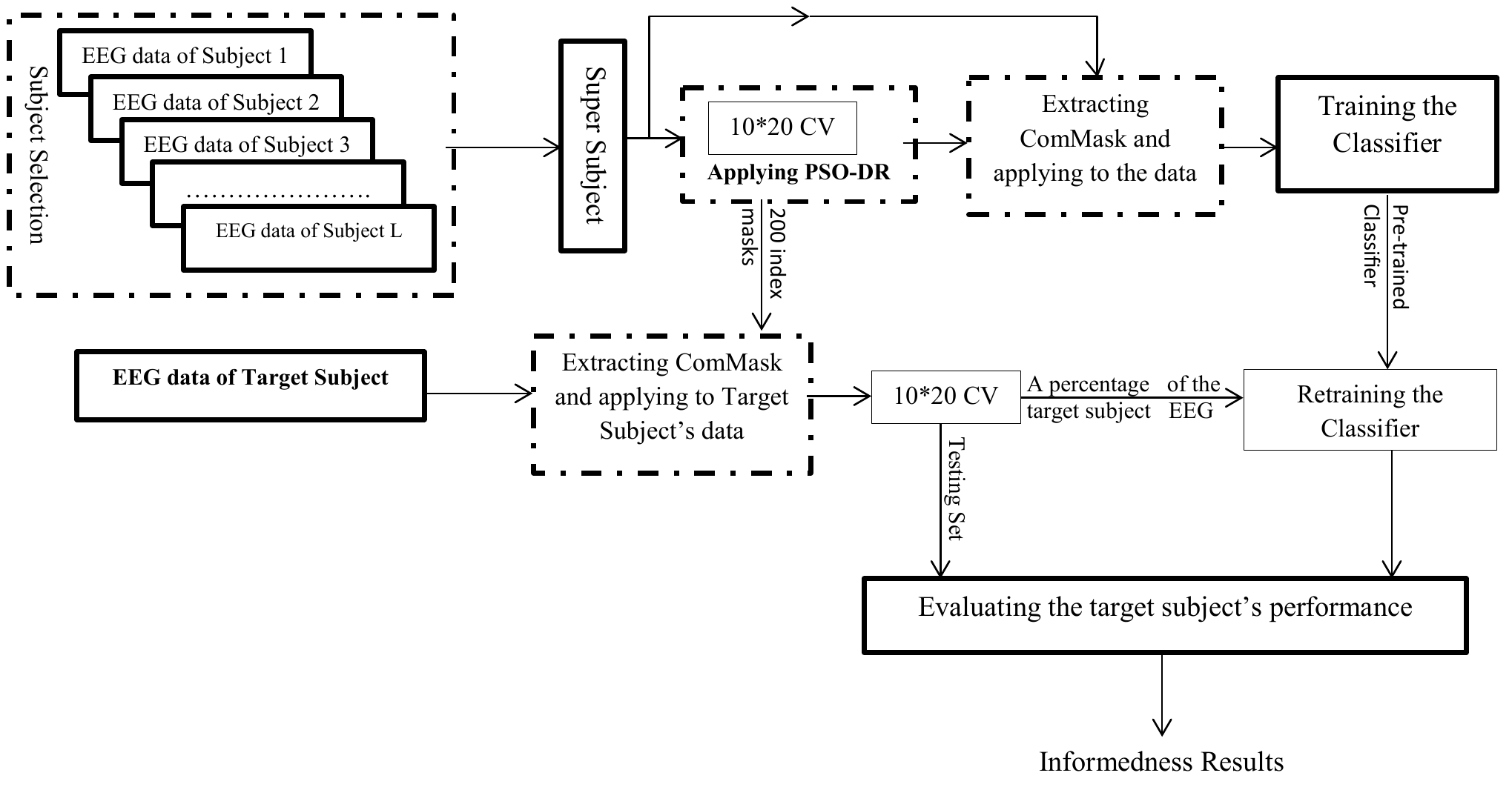}
\end{center}
\caption{A diagram demonstrating the process flow of Experiment 2. Dashed rectangles indicate optional steps that represent sub-experiments (referred to as conditions in the text).}
\label{Exp2}
\end{figure*}

Within each experiment, several conditions (sub-experiments) are designed based on following criterion:
\begin{itemize}
\item The experiments are replicated based on the application of the PSO-DR in order to study the impact of dimension reduction. 
\item The experiments are replicated based on the choice of the subjects to be involved in the creation of the super subject in subject transfer.
\end{itemize}
Figs \ref{Exp1}\ref{Exp2} illustrate exemplary diagrams of the discussed experiments.

The decision about the method to be utilized for subject transfer is based on the findings stating that the combination of \emph{Common Mask} and Framework 2 with super subjects that represent either groups of 4 subjects or \emph{Best possible combination of subjects} are likely to be the best choices for this dataset (dataset IVa from BCI2003 Competition) \cite{FRERPSOJournal}. 

\begin{landscape}
\begin{table*}
\begin{scriptsize}
\begin{center}
\caption{The informedness ($\pm$ Standard Error of mean) results of the best performing \emph{Super Subjects} with different targeted subjects using the combinations of ComMask, Framework 2 and Polynomial SVM and Perceptron with early stopping as classifiers\cite{FRERPSOJournal}.}
\begin{tabular}{c||c||cl|cl||c|c} 
\hline
\hline
&&\multicolumn{2}{c|}{Average} &\multicolumn{2}{c||}{Average}&&Notation \\
 Targeted&Training&&Accuracy (\%) \&&&Accuracy (\%) \&&Grouping&Utilized in\\
&&Informedness&Classification&Informedness&Classification&& Following\\
Subject&Subject/s&&Time (s)&&Time (s)&(G)&Experiments\\\cline{3-6}
&&\multicolumn{2}{c}{Polynomial SVM} &\multicolumn{2}{|c||}{Perceptron}&& \\
\hline
\hline
&AA&{0.36 $\pm$ 0.015}&68\% \& 344.45s& 0.28& 64\% \& 140.2282s&{FullSet}\\
&AL,AW,AY&{0.34 $\pm$ 0.016}&67\% \& 1.13s& 0.26 $\pm$ 0.027&63\% \&  62.7823s&G3&Bsub\\
AA&AL,AV,AW,AY&{0.35 $\pm$ 0.017}&68\% \& 2.12s& 0.20 $\pm$ 0.014 &60\% \& 70.5943s&G4&4sub\\
\hline
&AL&{0.72 $\pm$ 0.012}&86\% \& 317.14s& 0.67&83\% \& 115.0106s&{FullSet}\\
&AV,AW,AY&{0.67 $\pm$ 0.014}&83\% \& 0.58s& 0.56 $\pm$ 0.013&78\% \&  27.9410s&G3&Bsub\\
AL&AA,AV,AW,AY& 0.55 $\pm$ 0.018&78\% \& 1.39s& 0.43 $\pm$ 0.011&71\% \& 22.8348s&G4&4sub\\
\hline
&AV&0.22 $\pm$ 0.016&61\% \& 264.34s& 0.16&58\% \&  101.5296s& FullSet\\
&AA,AL,AW&{0.26 $\pm$ 0.020}&63\% \& 1.01s&0.15 $\pm$ 0.017&57\% \&  18.7323s&G3&Bsub\\
AV&AA,AL,AW,AY& 0.23 $\pm$ 0.019&61\% \& 2.33s& 0.17 $\pm$ 0.017&59\% \& 35.8243s&G4&4sub\\
\hline
&AW&{ 0.56 $\pm$ 0.013}&78\% \& 403.87s& 0.46&73\% \& 112.8258s&{ FullSet}\\
&AA,AY&{0.44 $\pm$ 0.017}&72\% \& 1.29s& 0.29 $\pm$  0.013&64\%  \& 31.3842s&G2&Bsub\\
AW&AA,AL,AV,AY& 0.39 $\pm$ 0.020&69\% \& 2.08s& 0.30 $\pm$ 0.015&65\% \& 19.9141s&G4&4sub\\
\hline
&AY&{0.51 $\pm$ 0.013}&75\% \& 265.49s&0.28&64\% \& 151.2197s&{FullSet}\\
&AL,AV,AW& 0.49 $\pm$ 0.018&74\% \& 0.93s& 0.40 $\pm$ 0.019& 70\% \& 20.4948s&G3&Bsub\\
AY&AA,AL,AV,AW&{0.50 $\pm$ 0.017}&75\% \& 2s&0.42 $\pm$ 0.025 &71\% \& 22.9320s&G4&4sub\\
\hline
\hline
\label{TABLE}
\end{tabular}
\end{center}
\end{scriptsize}
\end{table*}
\end{landscape}

Table \ref{TABLE} depicts the Informedness achieved with the best performing \emph{Super Subjects} using \emph{ComMask}, Framework 2 and Polynomial SVM. The number of subjects involved in the creation of the \emph{Super Subject} is denoted by a grouping factor (\emph{G$\in [2,3,4]$}). The results achieved without any dimension reduction, referred to as FullSet, are also included. These results are based on a 10 $\times$ 20 CV using 90\% of the data for training and the remaining 10\% for validation and testing (5\% each). Although the best results in \cite{FRERPSOJournal} are achieved by Polynomial SVM, since Perceptron is utilized as the primary classifier in this study, the results achieved with this classifier are also included in the table. 

Given that the aim of the study is to investigate the possibility of reducing the number of training samples originating from the target subject, varying percentages of the data originating from EEG of the target subject are utilized for training. To provide cross validation consistent with previous studies, a 10 $\times$ 20 CV is conducted resulting in three sets of indexes representing training, validation and testing among which the validation and the testing sets always represent 5\% of the data while the training set is varied based on the parametrization of the experiments (from 0\% representing Zero Trial EEG to 90\% representing Single Trial EEG with 5\% increase steps). The cross validation (CV) is performed in a way to prevent sub-epochs (0.5s) originating from the same super-epoch (2.5s) being appeared in more than one set (e.g., training, validation, or testing sets). In all experiments, a modified single layer Perceptron featuring early stopping is utilized for classification. The choice of using the  modified single layer Perceptron as the classifier in this study is made due to its feasibility to be utilized in two phases of training and retraining while the SVM (LIBSVM implementation \cite{CC01a}) does not facilitate this. This is with the understanding that in all previous studies Polynomial SVM showed better overall performance compared with the modified Perceptron \cite{WCCIEA,WCCIPSO, WCCICat, FRERPSOJournal}.

\subsection{Experiment 1: Without retraining}
In this set of experiments, various portions of the EEG data of the target subject are concatenated to the EEG data of the \emph{Super Subject} to be utilized to train the classifier. The \emph{Super Subject} is either generated as the result of concatenation of EEG signals of 4 subjects or the best possible combination of subjects. The remaining data from the target subject is utilized for validation and testing. The experiment is repeated with and without DR operation. PSO-DR explained in previous sections is utilized to provide 99\% dimension reduction. It should be noted that in the experimental condition in which DR is utilized, a \emph{ComMask} derived from applying the PSO-DR to the EEG signal of the \emph{Super Subject} (in a 10 $\times$ 20 CV paradigm\footnote{10 repetitions of 20 fold cross validation.}) is utilized, and the reduction operation is conducted on all training, testing and validation sets (See Fig \ref{Exp1} for more details).

In all experiments involving the use of a \emph{Super Subject}'s data or PSO-DR's reduction mask the choice of the subjects to participate in the creation of the \emph{Super Subject} and the derived PSO-DR mask is made based on the previous findings presented in Table \ref{TABLE}. A list of abbreviations and notations related to this experiment that are utilized in the following figures and discussions are presented in Table \ref{TResults1}.

\begin{table}
\begin{scriptsize}
\center
\caption{The notation utilized for representing the performed experiments and conditions.}
{\begin{tabular}{@{}c|c@{}} \hline\hline
{\bf Notation Utilized}&{\bf Explanation}\\\hline\hline
TS& Target Subject\\\hline
DR&Dimension Reduction\\\hline
${DR(y)}$& applying PSO-based Masks on \\
&subject (or Super Subject) \emph{y} resulting in 99\% DR\\
& The PSO mask utilized for DR is generated\\
& either on Bsub or 4sub in an off-line mode\\\hline
4sub& Super Subject with G=4\\\hline
Bsub& Best combination of subjects in a Super Subject\\\hline
Tr(x)& Initial Training on samples of subject (or Super Subject) \emph{x}\\\hline
Ret(y)& Retraining on samples of Target Subject \emph{y}\\\hline\hline
{\bf Experiment}&{\bf Notation Utilized}\\
{\bf Conditions}&{\bf }\\\hline\hline
{\bf Experiment 1.1}&\\\hline
Cond1.1a&$(TS)$\\
Cond1.1b&$(TS)+ (4sub)$\\
Cond1.1c&$(TS) + (Bsub)$\\\hline
{\bf Experiment 1.2}&\\\hline
Cond1.2a&${DR(TS)}$\\
Cond1.2b&${DR(TS)}+{DR(4sub)}$\\
Cond1.2c&${DR(TS)}$\\
Cond1.2d&${DR(TS)}+{DR(Bsub)}$\\\hline\hline
\end{tabular}}
\label{TResults1}
\end{scriptsize}
\end{table}

The experiments conducted are as follows:
\subsubsection{\bf \emph{Experiment 1.1: Using full-set without reduction - $1^{st}$ baseline set of approaches:}}
\begin{itemize}
\item Condition 1.1(a): Full-set without reduction and without extra training samples. This is denoted as \emph{$(TS)$} in figures.
\item Condition 1.1(b): Full-set without reduction and with extra training samples originating from a \emph{Super Subject} resulted from concatenation of 4 subjects' signal. This is denoted as \emph{$(TS) + (4sub)$} in figures.
\item Condition 1.1(c): Full-set without reduction and with extra training samples originating from a \emph{Super Subject} representing the best possible combination of subjects. This is denoted as \emph{$(TS) + (Bsub)$} in figures.
\end{itemize}

\begin{figure*}
\begin{center}
\includegraphics[width=5.5in]{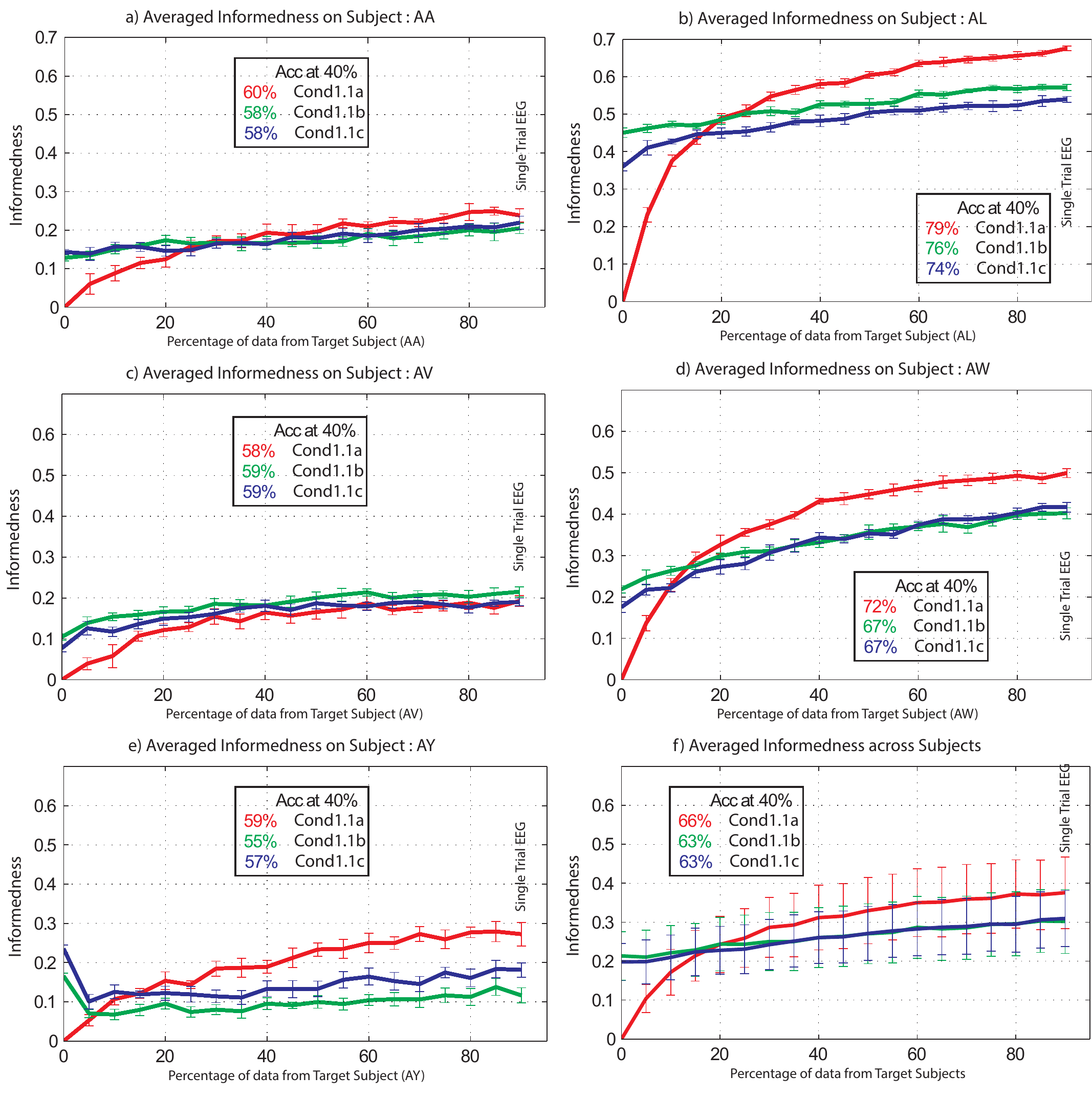}
\end{center}
\caption{Utility of various portions of EEG data for training within Condition 1.1 paradigm (without any dimension reduction operation). Cond1.1a indicate \emph{$(TS)$}, Cond1.1b  indicate \emph{$(TS) + (4sub)$}, and Cond1.1c  indicate \emph{$(TS) + (Bsub)$}. The error bars are standard error. The average of the classification accuracy for each condition when only 40\% of data originating from the target subject is utilized is denoted by Acc at 40\%.}
\label{Experiment1-1}
\end{figure*}
The impact of using various portions of training samples from the target subject in addition to the use of extra training samples from the \emph{Super Subject} (based on nominated individuals performing the same task) is demonstrated in Fig. \ref{Experiment1-1}. The results indicate that in case of using smaller portions (in the range of 0\% to 40\%) of the target subject's training samples, methods that import extra training samples originating from other subjects performing similar tasks achieved better averaged classification when using lower percentages of the training samples (up to 20\%). This issue is especially visible within the Zero Trial experiment (0\%). Given the ratio between the gained informedness and the percentage of  the training samples of the target subjects utilized, approximately 40\% of the target subject's data is sufficient for training the classifier (with respect to the lost informedness).

From Fig. \ref{Experiment1-1} it would appear that AL is performing consistently well and is already very competent at the task.  Thus the \emph{Super subject} data is enough to launch the learner to a high initial performance level, but later holds it back as the system tries to accommodate the error and inconsistency of the other subjects whose data is included - the linearity of these 1.1b and 1.1c curves indicates this is the only significant factor after the 10\% point for both.  A condition where \emph{Super Subject} data is reduced as target data becomes available should achieve the best of both worlds.

The best subject, AL, is included in the 4sub data for all other subjects, and the Bsub for all but AW, who we presume utilized a different strategy or belonged to a different EEG phenotype (and is the only one that borrowed two rather than three of the other subjects' data). Thus the consistency and expertise of AL benefits the other subjects, but they don't benefit AL nearly so much. Detailed information about the subjects that are involved in 4sub and Bsub of each target subject can be found in Table \ref{TABLE}. Generally, as noted in \cite{FRERPSOJournal}, the best combination (Bsub) consists of a mix of good and mediocre subjects, presumably as a good subject provides a good model, and a mediocre subject provides increased variance and noise immunity.

AW and AL are not as good, but the same effect is seen.  AA is a bit worse and the crossover lags. 

AV is the worst and is the only one that benefits from the other data until the very end.  This is likely to be caused by inconsistency due to being at an early stage of the learning curve, and there isn't any consistent strategy to tune into yet.  Note that AV is avoided as part of Bsub for all subjects except AY - possibly these subjects were at a similar stage in their learning curve, utilized a similar strategy, or had a similar EEG phenotype.

Considering the relationship between the Cond1.1b and 1.1c results, where Bsub includes helpful subjects and 4Sub is adding worse subjects (all subjects in the dataset except for the target subject), it is likely to see 4Sub performing worse, and this is seen for the best, and a strong affect of this nature for the poor subjects AA and AY is observed, where the worst subject AV is inflicted on them in Cond1.1b.

However, where there is no worse subject, and 4Sub is only adding better subjects, it is expectable to see 4Sub (1.1b) perform better, and this is visible for the worst subject AV.

In the case of the best subject AL, the broader model proves useful with the extra breadth outweighing the inconsistency of the weaker subjects (providing a balance between good model and good variance), and the effect is inconsistently damped as additional own data becomes available.  In the case of the second best subject AW, a similar but less pronounced effect is seen initially, but the difference is more thoroughly damped by additional own data, and AW is noteworthy for the fact that the best subject is not part of Bsub (1.1c) but is added in 4Sub (1.1b).
\subsubsection{\bf \emph{Experiment 1.2: 99\% DR through PSO-DR}}
\begin{itemize}
\item Condition 1.2(a):  99\% DR without extra training samples. The DR is performed using ComMask originating from a \emph{Super Subject} involving 4 subjects. This is denoted as \emph{${DR(TS)}$} in figures.
\item Condition 1.2(b): 99\% DR with extra training samples originating from a \emph{Super Subject} resulting from concatenation of 4 subjects' signals. This is denoted as \emph{${DR(TS)}+{DR(4sub)}$} in figures.
\item Condition 1.2(c):  99\% DR without extra training samples. The DR is performed using ComMask originating from a \emph{Super Subject} involving the best combination of subjects. This is denoted as \emph{${DR(TS)}$} in figures.
\item Condition 1.2(d): 99\% DR with extra training samples originating from a \emph{Super Subject} representing the best possible combination of subjects. This is denoted as \emph{${DR(TS)}+{DR(Bsub)}$} in figures.
\end{itemize}

\begin{figure*}
\begin{center}
\includegraphics[width=5.5in]{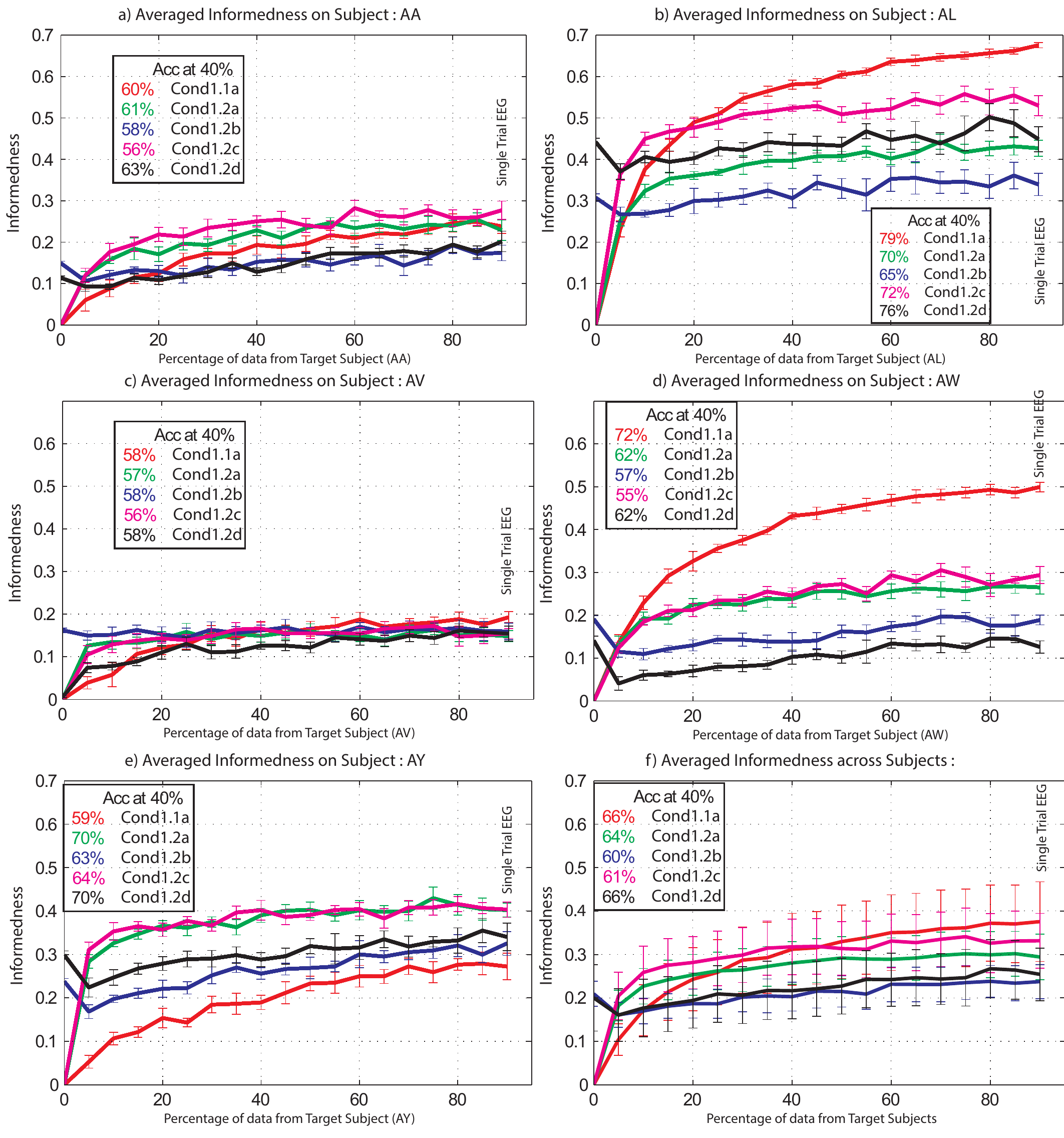}
\end{center}
\caption{Utility of various portions of EEG data for training within Condition 1.2 paradigm (the impact of 99\% dimension reduction). Cond1.1a represents \emph{$(TS)$}, Cond1.2a represents \emph{${DR(TS)}$}  using ComMask obtained from 4sub Super Subject for DR, Cond1.2b represents \emph{${DR(TS)}+{DR(4sub)}$},  Cond1.2c represents \emph{${DR(TS)}$} using ComMask obtained from Bsub Super Subject for DR, and Cond1.2d represents \emph{${DR(TS)}+{DR(Bsub)}$}. The error bars are standard error. The average of the classification accuracy for each condition when only 40\% of data originating from the target subject is utilized is denoted by Acc at 40\%.}
\label{Experiment1-2}
\end{figure*}

Fig. \ref{Experiment1-2} illustrates the achieved informedness with respect to the percentage of the data utilized from the target subjects. This experiment is a replication of Experiment 1.1 with the inclusion of PSO-DR resulting in 99\% reduction in the data dimensions. The results indicate the advantage for the poorer subjects of using extra training samples originating from other subjects. Given the ratio between the informedness gained and the percentage of other data utilized, it can be concluded that using smaller portions (as little as 40\%) of own training samples is sufficient for training the classifier and the inclusion of extra training samples of other subjects improves the classification performance in these subjects. Similar to Experiment 1.1 (Conditions 1.1 a,b, and c), the results achieved from the Zero Trial experiments (using 0\% of the data originating from the target subjects for training/calibrating the classifier) indicate the advantage of using training samples from groups of subjects that performed similar tasks (Cond1.2b and d), however this advantage is quickly dissipated (Cond1.2a and c show better performance with even 10\% own data) and it may also be noted that using a mask based on Bsub tends to lead to better performance than PSO-DR based on 4Sub (but is insignificantly or inconsistently different for  the weaker 3 subjects, where for the weakest (AV) none of the conditions gives significantly better or worse results than any of the others after 50\% own data is utilized for training, and for the two moderates (AA and AY) either DR mask results in significantly better performance than the other data masked and augmented conditions (1.2b and c) after even 10\% own data is trained on, and the Bsub mask gives better test results than even the unreduced data from this 10\% point right up to the 90\% point (significant to $p<0.05$ except for 3 of the 17 data points in this range for subject AA).  Thus the Bsub DR seems to give consistently better results than 4Sub DR across all subjects, and is never significantly worse than 4Sub DR for any subject.

In relation to the best two subjects (AL and AW), it is clear that the data reduction causes loss of generalized accuracy that is not made up for fully by any of the other data techniques utilized. Two hypotheses that may explain this are: 1. electrodes and features that are contributing to better performance are discarded due to the electrode reduction being too severe; and/or 2. they may have more overlapped brain activity with higher frequency complexity, so reduction of frequency bins is being too severe.

\subsection{Experiment 2: Retraining}
In this set of experiments, the EEG data of the \emph{Super Subject} is first utilized to train the classifier (in off-line mode) and later on, the classifier with the resulting weights is retrained and tested with varying portions of the EEG data of the target subject. The \emph{Super Subject} is either generated as the result of concatenation of EEG signals of 4 subjects or the best possible combination of subjects. Similar to experiment 1 the impact of PSO-DR is investigated using \emph{ComMask} originated from EEG data of the \emph{Super Subjects}  (See Fig \ref{Exp2} for more details).
The conducted experiments are as follows:
\subsubsection{\bf \emph{Experiment 2.1: Full-set without reduction- $2^{st}$ baseline set of approaches:}}

\begin{itemize}
\item Condition 2.1(a): Full-set without reduction; with classifiers trained on samples originating from a \emph{Super Subject} representing the concatenation of 4 subjects' signals; and later retrained on different portions of the target subject data. This is denoted as \emph{$Tr(4sub)+Ret(TS)$} in figures.
\item Condition 2.1(b): Full-set without reduction; the classifiers trained on samples originating from a \emph{Super Subject} representing the best possible combination of subjects; and later retrained on different portions of the target subject data. This is denoted as \emph{$Tr(Bsub)+Ret(TS)$} in figures.
\end{itemize}

\subsubsection{\bf \emph{Experiment 2.2: 99\% DR through PSO-DR}}

\begin{itemize}
\item Condition 2.2(a): 99\% DR; classifiers trained on samples originating from a \emph{Super Subject} resulting from concatenation of 4 subjects' signals; and later retrained on different portions of the target subject data. This is denoted as \emph{$Tr( {DR(4sub)})+Ret({DR(TS)})$} in figures.
\item Condition 2.2(b): 99\% DR; classifiers trained on samples originating from a \emph{Super Subject} representing the best possible combination of subjects; and later retrained on different portions of the target subject data. This is denoted as \emph{$Tr({DR(Bsub)})+Ret({DR(TS)})$} in figures.
\end{itemize}

 A list of abbreviations and notations related to this experiment that are utilized in the following figures and discussions are presented in Table \ref{TResults2}.
 
\begin{table}
\begin{scriptsize}
\center
\caption{The notation utilized for representing the performed experiments and conditions.}
{\begin{tabular}{@{}c|c@{}} 
\hline
\hline
{\bf Notation Utilized}&{\bf Explanation}\\\hline\hline
TS& Target Subject\\\hline
DR&Dimension Reduction\\\hline
${DR(y)}$& applying PSO-based Masks on \\
&subject (or Super Subject) \emph{y} resulting in 99\% DR\\
& The PSO mask utilized for DR is generated\\
& either on Bsub or 4sub in an off-line mode\\\hline
4sub& Super Subject with G=4\\\hline
Bsub& Best combination of subjects in a Super Subject\\\hline
Tr(x)& Initial Training on samples of subject (or Super Subject) \emph{x}\\\hline
Ret(y)& Retraining on samples of Target Subject \emph{y}\\\hline\hline
{\bf Experiment}&{\bf Notation Utilized}\\
{\bf Conditions}&{\bf }\\\hline\hline
{\bf Experiment 2}&\\\hline
Cond2.1a&$Tr(4sub)+Ret(TS)$\\
Cond2.1b&$Tr(Bsub)+Ret(TS)$\\
Cond2.2a&$Tr( {DR(4sub)})+Ret({DR(TS)})$\\
Cond2.2b&$Tr({DR(Bsub)})+Ret({DR(TS)})$\\\hline\hline
\end{tabular}}
\label{TResults2}
\end{scriptsize}
\end{table}

\begin{figure*}
\begin{center}
\includegraphics[width=5.5in]{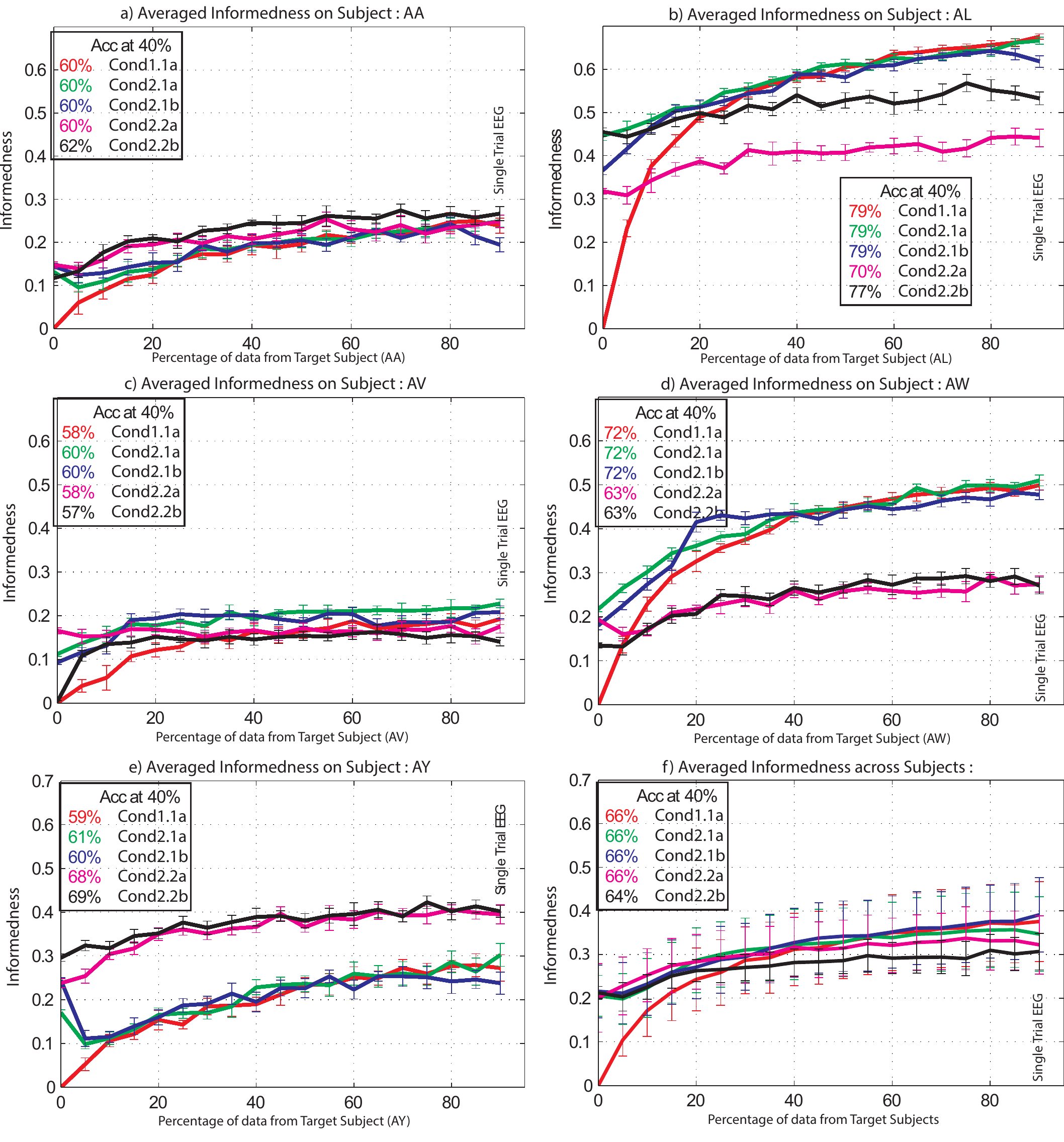}
\end{center}
\caption{The impact of retraining various portions of EEG data within Experiment 2 paradigm (with and without appliance of 99\% dimension reduction). Cond1.1a indicate \emph{$(TS)$}, Cond2.1a represents \emph{$Tr(4sub)+Ret(TS)$} and Cond2.1b represents \emph{$Tr(Bsub)+Ret(TS)$}. Cond2.2a and Cond2.2b experiments are similar to Cond2.1a and Cond2.1b  respectively with addition of PSO based 99\% DR (representing \emph{$Tr( {DR(4sub)})+Ret({DR(TS)})$} and \emph{$Tr({DR(Bsub)})+Ret({DR(TS)})$}). The error bars are standard error. The average of the classification accuracy for each condition when only 40\% of data originating from the target subject is utilized is denoted by Acc at 40\%.}
\label{Experiment2}
\end{figure*}

Experiment 2 is a replication of experiments 1.1 and 1.2 with the addition of an extra training step. The idea is to initially train the classifier on training samples of other subjects that performed similar tasks and later on further train the classifier using portions of the data originating from the target subjects. This procedure is performed with (condition 2.2) and without  (condition 2.1) appliance of the PSO-DR operator. The achieved results are demonstrated in Fig. \ref{Experiment2}. The results show no significant and clear difference between the conditions that incorporated DR operator (Cond2.2a and b) and those that did not (Cond2.1a and b) in poorer subjects except with subject AY in which the appliance of the PSO-DR operator clearly improved the informedness with no clear difference between the use of 4sub or Bsub as the Super Subject.  Given the ratio between the informedness gained and the percentage of others' data utilized, it can be concluded that using smaller portions (as little as 40\%) of own training samples is sufficient for training the classifier and the inclusion of extra training samples of other subjects improve the classification performance in these subjects. The results achieved from the Zero trial experiment (using 0\% of the own data for training/calibrating the classifier) indicate the advantage of using training samples coming from groups of other subjects that performed similar task. This issue is specially visible in subject AY within Cond2.1a and b in which the inclusion of own training samples degraded the informedness when using less than 20\% of own data for Cond2.1a and 55\% for Cond2.1b.  Similar to Cond1.2a and c, in the best two subjects (AL and AW), the data reduction causes the loss of informedness. This issue is more severe in subject AW while the combination of data reduction and pre-training model using Bsub as super subject is competitive in lower percentage ranges of 0\% to 30\% and shows only 0.1 informedness loss in higher ranges of 50\% to 90\%.

Similar to previous conditions, 40\% of the training samples originating from the target subjects is likely to provide reasonable ratio between the gained informedness and the percentage of target subject's data utilized with small lost of informedness in compared to the use of higher percentages. 

A comparison among the results illustrated in Figs \ref{Experiment1-1}, \ref{Experiment1-2}, and \ref{Experiment2} indicate the feasibility of the proposed PSO-DR operator within conditions that use lower percentages of data originating from target  subjects while the best overall informedness achieved through the use of higher portions of target subjects' data with consideration of 40\% as the threshold point after which the ratio between the achieved informedness and the percentage of own data utilized is no longer significant for higher amounts.

\begin{figure*}
\begin{center}
\includegraphics[width=5.5in]{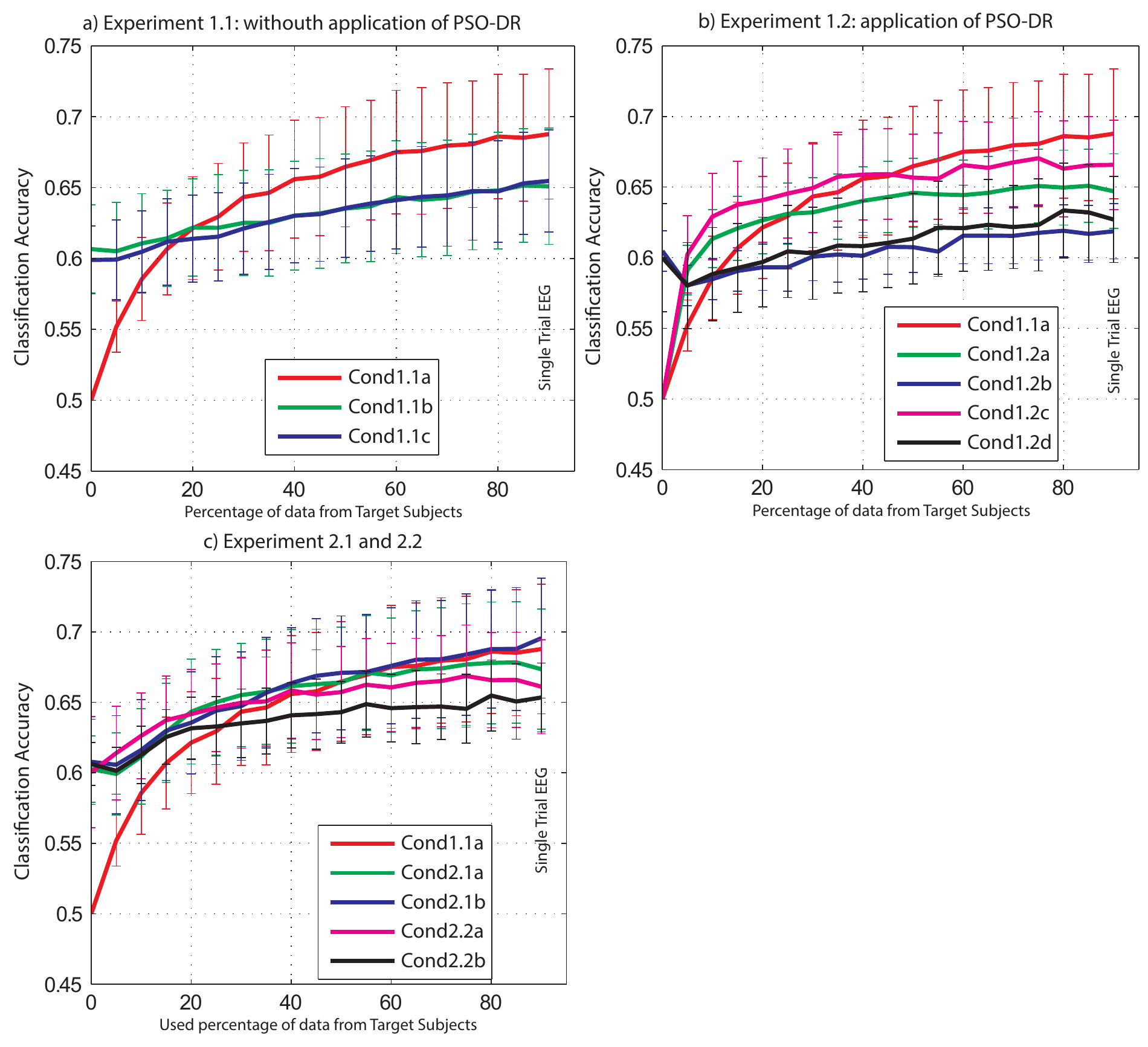}
\end{center}
\caption{The average classification accuracy across subjects using different training approaches and various percentages of data originating from the target subject. The error bars are standard error.}
\label{ExperimentAcc}
\end{figure*}

To provide better understanding of the results, the average of classification accuracy is presented in Fig \ref{ExperimentAcc}. The graphs in the figure represent the average of classification accuracy across subjects.

\subsection{Main Effects}
\label{Results}
Following main effects are considered in this study.
\begin{enumerate}
\item the impact of using different percentages of data for training
\item the impact of subjects  on achieved performance
\item the impact of retraining
\item the impact of dimension reduction
\item the interaction of the last two main factors
\end{enumerate}

\begin{figure*}
\begin{center}
\includegraphics[width=5.5in]{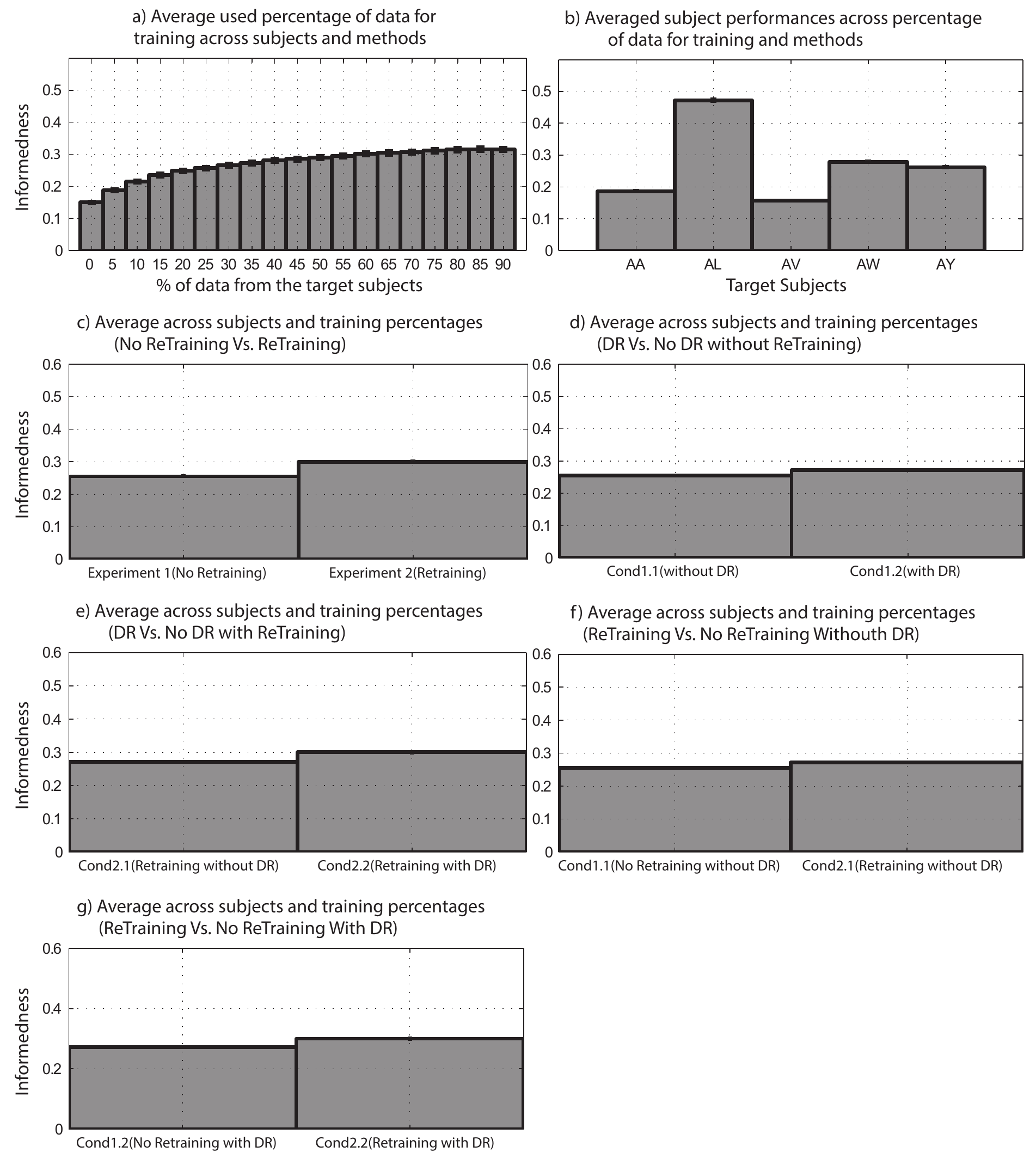}
\end{center}
\caption{Demonstration of the impact of 5 main effects (Utilized percentage for training, Subject, Retraining, Dimension Reduction (DR), and using Extra training samples of other subjects) and 3 selected interactions. The selected interactions focuses on the cross relations between DR(Cond1.1 vs 1.2 and Cond2.1 vs 2.2), retraining (Exp1 vs Exp2), and having extra training sample regardless of their origin (super subject either containing the best combination of subjects or as many subject left as possible (4 remaining subjects in this dataset)). The error bars are standard error.}
\label{MainEffect}
\end{figure*}

These principal factors and their suggested interactions are demonstrated in Fig. \ref{MainEffect}. The figure presents the average results across different factors such as utilized percentage of data, subjects, experiments, etc. The following conclusions can be made regarding the factors considered:
\begin{enumerate}
\item \emph{The required Percentage from the target subject for training}: The results of using between 0\% to 90\% of data for training the classifier indicates a significant difference between 0\% and 90\% with 40\% being the threshold for 0.05 significance of the loss of Informedness (viz. $p < 0.05$ in the range [0\%,40\%] and $p > 0.05$ in the range [40\%, 90\%]). The use of only 40\% of recorded EEG for training is equivalent to using 560 (out of 1400) sub-epochs of 0.5s length which reduces the amount of required EEG data for calibration and training to less than 5 min. Of course significance is itself dependent on the power of the experimental paradigm and the amount of data available, so that what is more important is the apparent linearization of the asymptote from around 40\%, but it appears continues to arise until 85\%, but with increasingly small gains.
\item \emph{The impact of subjects}: Among 5 subjects in the dataset, subject \emph{AL} shows significantly better performance than the others ($p< 0.05$).
\item \emph{The impact of retraining}: The results shows better overall performance for experiments that incorporate retraining (denoted as Exp 2 that contains Condition 2.1 (a and b) and Condition 2.2 (a and b)) with a significant difference when the extra training step is performed ($p< 0.05$).
\item \emph{Dimension Reduction (DR)}: 
Dimension reduction shows slightly better Informedness performance on average in comparison with experimental conditions that uses the full set of data without application of PSO-DR. This feature is illustrated through two interaction between the use of PSO-DR with and without retraining. In figure, Cond2.1 and Cond1.1 refers to the use of full set with no PSO-DR application with and without retraining respectively. Cond2.2 and Cond1.2 refers to the use of DR with and without retraining. This conclusion is made based on the illustrated results of interactions between Cond1.1 and Cond1.2 and also Cond2.1 and Cond2.2. It is noteworthy that due to 99\% reduction caused by the PSO-DR method, the required time for training on average is reduced to less than 2s which makes the outcome desirable for on-line systems.
\end{enumerate}

\subsection{Classification Time}
\begin{figure*}
\begin{center}
\includegraphics[width=5.5in]{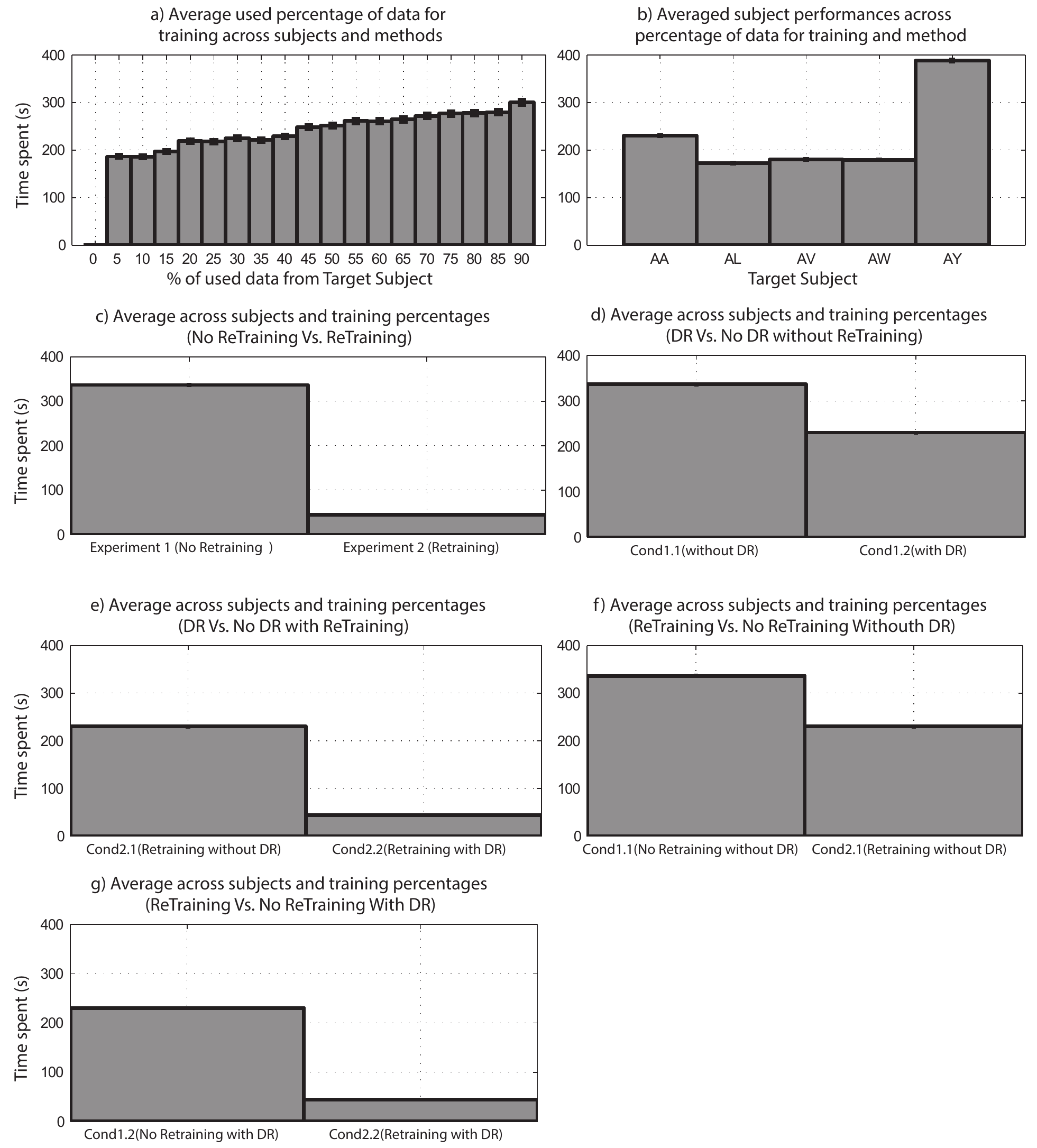}
\end{center}
\caption{{\bf Demonstration of the impact of 5 main effects} (Utilized percentage for training, Subject, Retraining, Dimension Reduction (DR), and using extra training samples of other subjects) and 3 selected interactions. The selected interactions focuses on the cross relations between DR(Cond 1.1 vs 1.2 and 2.1 vs 2.2), retraining (Exp1 vs Exp2), and having extra training sample regardless of their origin (super subject either containing the best combination of subjects or as many subject left as possible). The error bars are standard error.}
\label{MainEffectTime}
\end{figure*}

\begin{figure*}
\begin{center}
\includegraphics[width=5.5in]{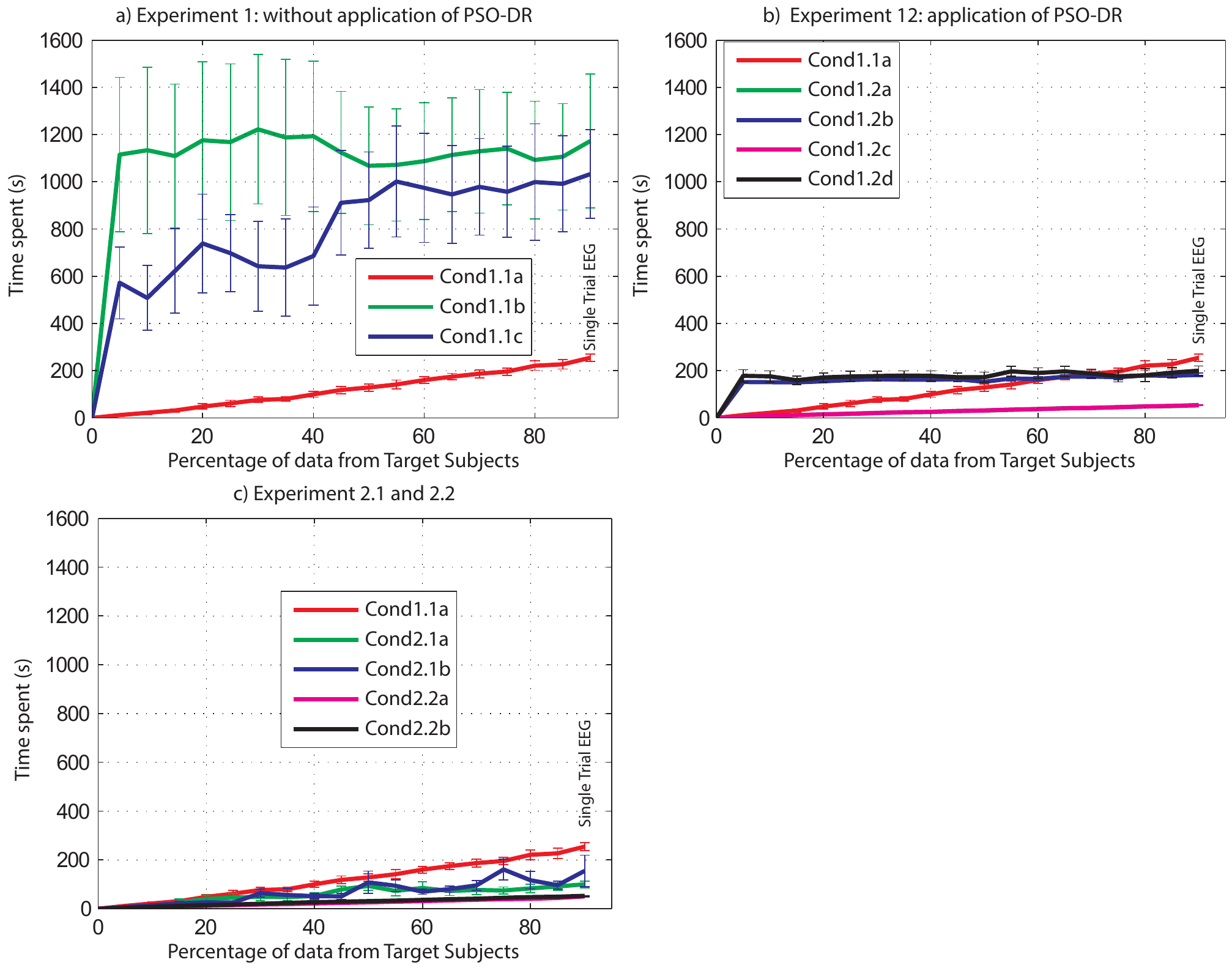}
\end{center}
\caption{{\bf The average time spent} across subjects using different training approaches and varying percentage of target subject data. The error bars are standard error.}
\label{ExperimentTime}
\end{figure*}

Figs \ref{MainEffectTime} and \ref{ExperimentTime} depicts the average time spent for training using various portions of the target subjects data for training. The results indicate that the inclusion of extra training samples of other subjects in all cases increases the required classification training time (across all variations of experiment 1). The use of two stages of training (off-line mode) and retraining (on-line mode) stages in experiment 2 is likely to be a solution for this issue. The results illustrate lower amount of time spent for the second training stage (retraining stage) in experiment 2 in comparison with only using target subjects data for training (Cond1.1a). The application of the PSO-DR operator in Cond2.2a and b decreases the required retraining time. The inclusion of higher percentages of the target subject's data increases the required training time but the difference is not significant for experiment 2 specially when PSO-DR operator is applied ($p>0.05$). The results in Fig. \ref{MainEffectTime} indicate that among subjects, subject AY's data required the highest amount of time on average for training the classifier. Looking at the main effect interactions, the application of PSO-DR operator reduced the required classification time specially when utilized within two stages of training and retraining (experiment 2). The reported average spent time in Figs \ref{MainEffectTime} and \ref{ExperimentTime} is with the understanding that the experiments were executed on a diverse range of machines. 

\subsection{Overall Achievements}
The average of the achieved performance with various subjects using the methodologies discussed in this study for reducing the dependency to targeted subject's data is depicted in Table \ref{T:chap75}. The first set of results for each subject represents the situation that only targeted subject's data is utilized (the performance baseline). Following issues are noteworthy:
\begin{itemize}
\item The observed degradation of performance with zero trial experiments (using 0\% of the target subject\rq{}s data for training the classifier) are consistent with observed outcomes by Krauledat et al., \cite{nosubjecttraining7} when target subject\rq{}s previous recording sessions\rq{} data were utilized for training the classifiers and Fazli et al. \cite{nosubjecttraining4} findings when other subjects' data is utilized for pre-training the classifiers.
\item Combination of Dimension Reduction (DR) and pre-trained classifiers showed potential on all variation of subjects with different EEG expertise (weak (AA, AV), normal (AW, AY), strong (AL)). This is an important achievement that allow the methodology to be useful for all variations of subjects. That is, a novice user with weak EEG expertise can improve his/her performance using the suggested methodology and after becoming more experienced (through conducting several practice sessions) the methodology would still be useful.   
\item Although the best achieved performance when only 40\% of the target subject's data is utilized usually achieved with pre-trained classifiers and without the application of PSO-DR, the applied dimension reduction shortened the required classifier's training.
\end{itemize}

\begin{table*}
\center
\begin{scriptsize}
\caption{The average of the achieved Informedness and the classifier training time for the combination of ComMask, Framework2, and Perceptron in on-line mode. Three categories of zero trial, 40\% of target subject,  and single trial (90\%) are reported.}
\begin{tabular}{|c|ccc||ccc|} 
\hline
{\bf Targeted}&\multicolumn{6}{|c|}{\bf Averaged achieved performance in on-line mode}\\\cline{2-7}
{\bf Subject}&{\bf Zero}&{\bf 40\% of Target}&{\bf Single Trial}&{\bf Zero}&{\bf 40\% of Target}&{\bf Single Trial}\\
{\bf (TS)}&{\bf Trial}&{\bf Subject (TS)}&{\bf (90\% TS)}&{\bf Trial}&{\bf Subject (TS)}&{\bf (90\% TS)}\\
\hline\hline
AA&0 \& 0.00s&0.21 \& 122.87s&\multicolumn{1}{c}{0.25 \& 282.90s}&\multicolumn{3}{c|}{\bf without DR without subject transfer}\\\cline{2-7}
&\multicolumn{6}{|c|}{\bf without retraining}\\
&\multicolumn{3}{|c||}{\bf without DR with extra training samples}&\multicolumn{3}{|c|}{\bf with DR \& no extra training samples}\\
&0.12 \& 0.00s&0.28 \& 1140.28s&0.21 \& 1151.26s&0 \& 0.00s&0.22 \& 24.70s&0.21 \& 54.28s\\
&&&&\multicolumn{3}{|c|}{\bf with DR \& with extra training samples}\\
&&&&0.15 \& 0.00s&0.16 \& 131.45s&0.19 \& 204.68s\\
&\multicolumn{6}{|c|}{\bf with retraining}\\
&\multicolumn{3}{|c||}{\bf without DR with extra training samples}&\multicolumn{3}{|c|}{\bf with DR \& with extra training samples}\\
&0.13  \& 0.00s&0.2 \& 30.88s&0.25 \& 162.67s&0.14 \& 0.00s&0.2 \& 28.48s&0.24 \& 50.50s\\
\hline
\hline
AL& \& 0.00s&0.6 \& 71.69s&\multicolumn{1}{c}{0.68 \& 228.44s}&\multicolumn{3}{c|}{\bf without DR without subject transfer}\\\cline{2-7}
&\multicolumn{6}{|c|}{\bf without retraining}\\
&\multicolumn{3}{|c||}{\bf without DR with extra training samples}&\multicolumn{3}{|c|}{\bf with DR \& no extra training samples}\\
&0.46 \& 0.00s&0.52 \& 728.69s&0.59 \& 800.12s&0 \& 0.00s&0.4 \& 27.41s&0.43 \& 55.79s\\
&&&&\multicolumn{3}{|c|}{\bf with DR \& with extra training samples}\\
&&&&0.31 \& 0.00s&0.31 \& 168.92s&0.33 \& 194.61s\\
&\multicolumn{6}{|c|}{\bf with retraining}\\
&\multicolumn{3}{|c||}{\bf without DR with extra training samples}&\multicolumn{3}{|c|}{\bf with DR \& with extra training samples}\\
&0.45 \& 0.00s&0.60 \& 35.76s&0.68 \& 81.76s&0.31 \& 0.00s&0.42 \& 25.14s&0.43 \& 52.55s\\
\hline
\hline
AV&0 \& 0.00s&0.15 \& 70.34s&\multicolumn{1}{c}{0.2 \& 210.06s}&\multicolumn{3}{c|}{\bf without DR without subject transfer}\\\cline{2-7}
&\multicolumn{6}{|c|}{\bf without retraining}\\
&\multicolumn{3}{|c||}{\bf without DR with extra training samples}&\multicolumn{3}{|c|}{\bf with DR \& no extra training samples}\\
&0.1 \& 0.00s&0.19 \& 798.44s&0.21 \& 780.52s&0 \& 0.00s&0.15 \& 24.25s&0.14 \& 53.32s\\
&&&&\multicolumn{3}{|c|}{\bf with DR \& with extra training samples}\\
&&&&0.17 \& 0.00s&0.16 \& 171.67s&0.15 \& 176.93s\\
&\multicolumn{6}{|c|}{\bf with retraining}\\
&\multicolumn{3}{|c||}{\bf without DR with extra training samples}&\multicolumn{3}{|c|}{\bf with DR \& with extra training samples}\\
&0.11 \& 0.00s&0.19 \& 28.72s&0.21 \& 45.51s&0.16 \& 0.00s&0.15 \& 23.19s&0.16 \& 47.62s\\
\hline
\hline
AW&0 \&0.00s&0.43 \& 105.03s&\multicolumn{1}{c}{0.5 \& 256.30s}&\multicolumn{3}{c|}{\bf without DR without subject transfer}\\\cline{2-7}
&\multicolumn{6}{|c|}{\bf without retraining}\\
&\multicolumn{3}{|c||}{\bf without DR with extra training samples}&\multicolumn{3}{|c|}{\bf with DR \& no extra training samples}\\
&0.21 \& 0.00s&0.33 \& 858.12s&0.41 \& 856.69s&0 \& 0.00s&0.24 \& 27.53s&0.27 \& 56.23s\\
&&&&\multicolumn{3}{|c|}{\bf with DR \& with extra training samples}\\
&&&&0.2 \& 0.00s&0.14 \& 182.33s&0.2 \& 176.73s\\
&\multicolumn{6}{|c|}{\bf with retraining}\\
&\multicolumn{3}{|c||}{\bf without DR with extra training samples}&\multicolumn{3}{|c|}{\bf with DR \& with extra training samples}\\
&0.21 \& 0.00s&0.42 \& 83.12s&0.52 \& 83.11s&0.2 \& 0.00s&0.26 \& 29.85s&0.28 \& 58.12s\\
\hline
\hline
AY&0 \& 0.00s&0.19 \& 127.78s&\multicolumn{1}{c}{0.28 \& 294.61s}&\multicolumn{3}{c|}{\bf without DR without subject transfer}\\\cline{2-7}
&\multicolumn{6}{|c|}{\bf without retraining}\\
&\multicolumn{3}{|c||}{\bf without DR with extra training samples}&\multicolumn{3}{|c|}{\bf with DR \& no extra training samples}\\
&0.18 \& 0.00s&0.1 \& 2436.91s&0.11 \& 2274.07s&0 \& 0.00s&0.39 \& 22.90s&0.4 \& 50.36s\\
&&&&\multicolumn{3}{|c|}{\it \bf with DR\& with extra training samples}\\
&&&&0.23 \& 0.00s&0.25 \& 152.70s&0.31 \& 157.22s\\
&\multicolumn{6}{|c|}{\bf with retraining}\\
&\multicolumn{3}{|c||}{\bf without DR with extra training samples}&\multicolumn{3}{|c|}{\bf with DR \& with extra training samples}\\
&0.16 \& 0.00s&0.22 \& 77.07s&0.3 \& 400.87s&0.24 \& 0.00s&0.37 \& 27.30s&0.4 \& 52.78s\\
\hline
\end{tabular}
\label{T:chap75}
\end{scriptsize}
\end{table*}
\section{Conclusion}
\label{Conclusion}
This study focused on providing comparison among the proposed methodologies that takes advantage from samples originating from groups of subjects that performed similar tasks in previous sessions. The selection criteria for including subjects to the set is made based on the previous findings in \cite{WCCICat} and \cite{FRERPSOJournal}. In addition, the impact of dimension reduction through the use of a PSO-based approach that reduces 99\% of the EEG data dimensions is investigated. Finally, a set of conditions that include an extra training step are evaluated. The main aim of the study was to generate a methodology that allows the inclusion of extra training samples from other subjects while reducing the dimensions of the data. Advantages like i) having extra training samples originating from other subjects recorded and pre-processed in an off-line mode, ii) minimum dependency to training samples from the target subject (the subject that performs the tasks in on-line mode), and iii) low dimensionality of the data caused by the applied evolutionary dimension reduction approach that minimizes the required classifier's training period can be considered as the main points of attraction for such a system in on-line EEG based BCI systems. 

Aiming to investigate the possibility of reducing the required own training samples, various percentages of the data originating from EEG of the target subject was utilized for training (from 0\% to 90\% in 5\% steps). The results suggested 40\% as the threshold for the amount of own (target subject) data required for training above which the ratio between the gained informedness and the amount of data utilized showed no significant improvement. This is equivalent to using 560 out of 1400 of 0.5s sub-epochs (less than 5 min EEG recordings) for training the classifier. Among 5 subjects in the dataset, subject \emph{AL} showed significant difference from others. The results indicate better overall performance for conditions that incorporate retraining. Dimension reduction through the use of PSO-DR operator showed slightly better performance on average in comparison with conditions that uses the full set of data with no DR. This is noteworthy since the employed DR method reduced 99\% of data which on average resulted in spending less than 10s for training the classifier. The results indicate the feasibility of the proposed PSO-DR operator within conditions that utilized lower percentages of data originating from target  subjects (between 0\% to 40\%) within inclusion of samples originating from other subjects that performed similar tasks (in an off-line mode). This effect is more visible within weaker subjects, especially in schemes that first trained the classifiers using others subjects' data (in an off-line mode) and then retrained the classifiers using lower percentages of own (target subject) data.

It is important to do not over interpret these results. It should be noted that the EEG data utilized in this study are originated from healthy subjects and even though interesting and encouraging results are achieved with weaker subjects, this can not be interpreted as an indication of suitability of the proposed methodology for paralyzed patients, specially those who have been paralyzed for a long time.

\section*{References}

\bibliography{PlosOne}

\end{document}